\documentclass[11pt]{article}

\usepackage[preprint]{acl}

\usepackage{times}
\usepackage{latexsym}
\usepackage{tabularx}
\usepackage[T1]{fontenc}

\usepackage[utf8]{inputenc}

\usepackage{microtype}

\usepackage{inconsolata}

\usepackage{graphicx}

\usepackage{booktabs}
\usepackage{todonotes}
\usepackage{multirow}
\usepackage{makecell}
\usepackage{enumitem}
\usepackage{svg}
\usepackage[most]{tcolorbox}
\usepackage{xcolor}

\definecolor{promptgreen}{RGB}{20,150,105}
\definecolor{promptgreenbg}{RGB}{238,250,246}

\definecolor{promptorange}{RGB}{205,92,10}
\definecolor{promptorangebg}{RGB}{255,247,240}

\definecolor{promptblue}{RGB}{80,95,120}
\definecolor{promptbluebg}{RGB}{247,248,250}

\newtcolorbox{greenpromptbox}[1]{
  enhanced,
  breakable,
  colback=promptgreenbg,
  colframe=promptgreen,
  colbacktitle=promptgreen,
  coltitle=white,
  title=#1,
  fonttitle=\bfseries,
  fontupper=\small\ttfamily,
  boxrule=0.6pt,
  arc=3pt,
  left=6pt,
  right=6pt,
  top=6pt,
  bottom=6pt,
  before skip=6pt,
  after skip=8pt
}

\newtcolorbox{orangepromptbox}[1]{
  enhanced,
  breakable,
  colback=promptorangebg,
  colframe=promptorange,
  colbacktitle=promptorange,
  coltitle=white,
  title=#1,
  fonttitle=\bfseries,
  fontupper=\small\ttfamily,
  boxrule=0.6pt,
  arc=3pt,
  left=6pt,
  right=6pt,
  top=6pt,
  bottom=6pt,
  before skip=6pt,
  after skip=8pt
}

\newtcolorbox{bluepromptbox}[1]{
  enhanced,
  breakable,
  colback=promptbluebg,
  colframe=promptblue,
  colbacktitle=promptblue,
  coltitle=white,
  title=#1,
  fonttitle=\bfseries,
  fontupper=\small\ttfamily,
  boxrule=0.6pt,
  arc=3pt,
  left=6pt,
  right=6pt,
  top=6pt,
  bottom=6pt,
  before skip=6pt,
  after skip=8pt
}
\usepackage{fontawesome5}

\title{Reasoning that Travels: Dissecting How Chain-of-Thought \\ Transfers Across Models}

\author{
\textbf{Xinyuan Cheng\thanks{\; Equal contribution.}\textsuperscript{\faMountain}} \quad
\textbf{Beiduo Chen\footnotemark[1]\textsuperscript{\faMountain\kern1pt\faRobot}} \quad
\textbf{Philipp Mondorf\textsuperscript{\faMountain\kern1pt\faRobot}} \quad
\textbf{Barbara Plank\textsuperscript{\faMountain\kern1pt\faRobot}}
\\[8pt]
\textsuperscript{\faMountain}MaiNLP, Center for Information and Language Processing, LMU Munich, Germany \\
\textsuperscript{\faRobot}Munich Center for Machine Learning, Germany \\
{\tt{ \href{mailto:xinyuan.cheng@campus.lmu.de}{\textcolor{black}{xinyuan.cheng@campus.lmu.de}}, \{\href{mailto:beiduo.chen@lmu.de}{\textcolor{black}{beiduo.chen}}, \href{mailto:p.mondorf@lmu.de}{\textcolor{black}{p.mondorf}}, \href{mailto:b.plank@lmu.de}{\textcolor{black}{b.plank}}\}@lmu.de}}
}

\begin{document}
\maketitle
\begin{abstract}

Large reasoning models (LRMs) often generate extensive chain-of-thought (CoT) traces before producing a final answer. As explicit textual artifacts, these traces can be passed to other models to solve the same task, enabling \emph{cross-model reasoning transfer}. Yet successful transfer alone does not reveal how the provided CoT contributes to another model's answer.
We study this question with a controlled provider--receiver framework, where a \textit{provider} generates a reasoning trace and a \textit{receiver} solves the same problem from increasingly longer trace prefixes. We compare \textit{force-answer}, where the receiver answers directly from the prefix, with \textit{free-generation}, where it may continue reasoning before answering.
Across models and benchmarks, full traces often transfer successfully, but prefix trajectories reveal distinct mechanisms. In force-answer mode, AIME transfer is largely driven by explicit answer availability. MMLU-Pro instead reflects a larger role for receiver competence, while ZebraLogic depends on partial structured-answer information rather than complete-answer leakage alone. In free-generation mode, partial CoTs improve performance across benchmarks, indicating that prefixes can guide continued reasoning. Finally, answer agreement among receivers provides a gold-free signal for stopping provider reasoning early.
Overall, cross-model CoT transfer is not a single phenomenon: it can reflect answer extraction, reasoning scaffolding, or receiver-dependent competence.

\end{abstract}

\section{Introduction}

\begin{figure*}[t]
    \centering
    \includegraphics[width=0.95\textwidth]{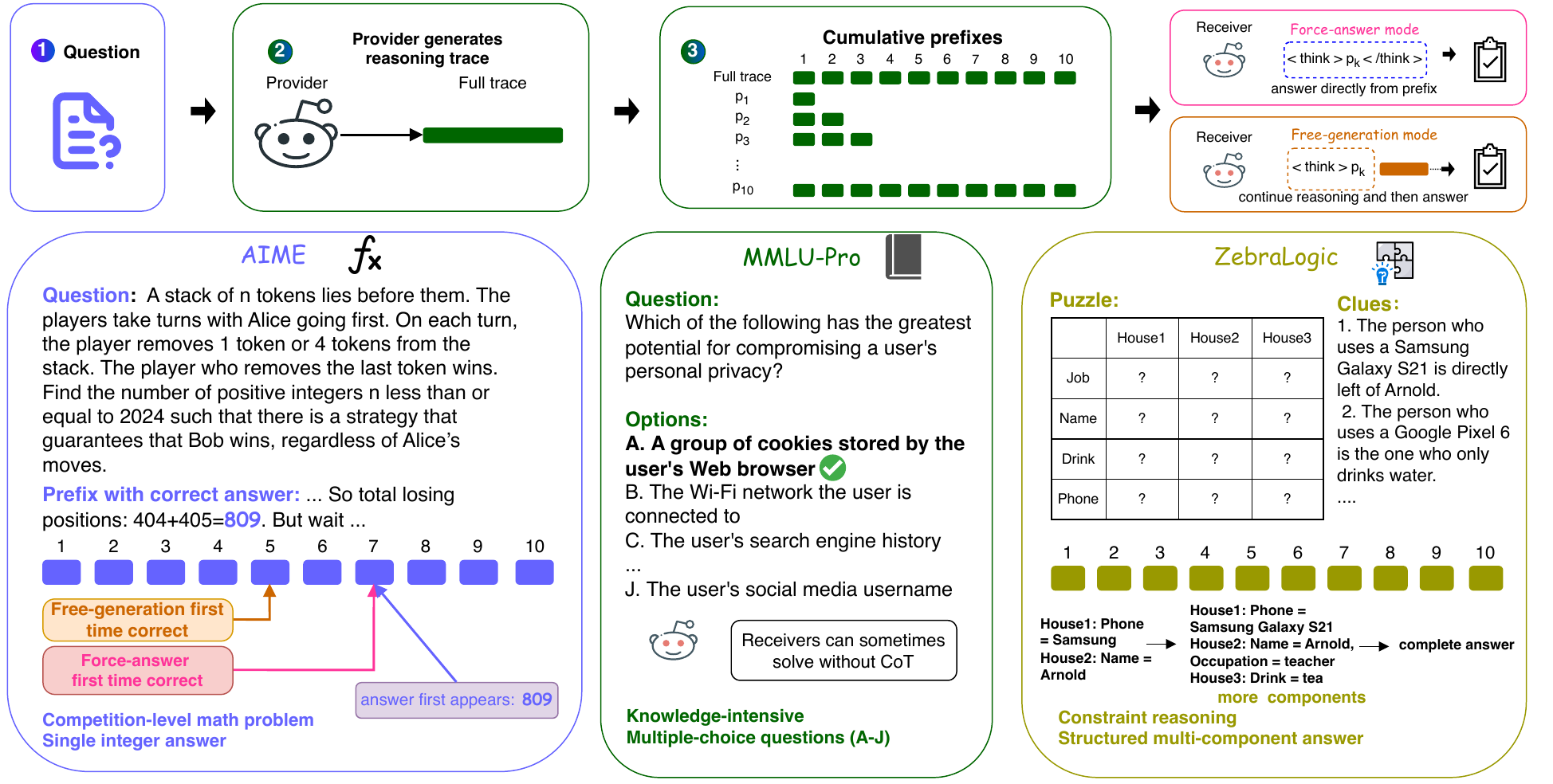}
\caption{
Overview of our provider--receiver reasoning-transfer framework.
Receivers are conditioned on cumulative portions of a provider trace and solve the same problem either by force-answering or free-generation.
The lower panels show how the framework is instantiated across benchmarks with different answer and reasoning structures.
}
    \label{fig:overview}
\end{figure*}

Large reasoning models (LRMs) achieve strong performance on complex multi-step tasks by generating long chain-of-thought (CoT) traces before final answers~\cite{gpto1, deepseekr1, olmo2026olmo3, bakouch2025smollm3, qwen3technicalreport}. Although such traces need not faithfully reflect the generating model's internal computation~\cite{paul-etal-2024-making, lanham2023measuringfaithfulnesschainofthoughtreasoning, turpin2023faithful,mondorf-plank-2024-comparing,arcuschin2025chainofthought,chen2025reasoning}, they may still serve as useful textual artifacts that convey task-relevant information~\cite{mondorf2024beyond}. Recent work operationalizes this view in cross-model settings, where traces from one model are reused, verified, continued, or passed to other models for answer generation, verification, and correction~\cite{roytburg2026measuringreasoningtracelegibility, zhao2026trigreasontriggerbasedcollaborationsmall, shi-etal-2025-speccot, li2026offtrajectory, liu2026confspecefficientsteplevelspeculative}.

Yet successful cross-model use alone does not reveal how the trace helps. Prior work mostly asks whether such reuse improves final performance or inference efficiency~\cite{pal2026explanationsgeneralizelargereasoning, bi2025cotxadaptiveframeworkcrossmodel}. This leaves open \textit{what information in the CoT is actually used and how transfer emerges as the trace unfolds.} A trace may simply contain the answer, provide partial reasoning that helps another model continue, or interact with the receiving model's own competence. We therefore conduct a controlled provider--receiver study across multiple models and benchmarks: a \textit{provider} generates a reasoning trace for a problem, and a \textit{receiver} solves the same problem using that trace.

We make two design choices to expose the mechanism of transfer. First, to locate when a provider trace becomes sufficient, we evaluate not only the completed trace but also cumulative prefixes of the same reasoning process. This shows how receiver performance changes as more provider reasoning becomes available. Second, because reasoning-tuned receivers may not naturally answer from an abruptly truncated CoT, we compare two trace-use modes. In \textit{force-answer} mode, the receiver answers directly from the available prefix, isolating what can be extracted from the provided text. In \textit{free-generation} mode, it may continue reasoning before answering, testing whether partial prefixes scaffold the receiver's own reasoning. 

Figure~\ref{fig:overview} summarizes our provider--receiver reasoning-transfer framework and illustrates the benchmark-specific phenomena analyzed below: explicit answer availability on AIME~\cite{matharena}, receiver-intrinsic competence on MMLU-Pro~\cite{mmlupro}, and structured partial-answer accumulation on ZebraLogic~\cite{zebralogic}.
Across three benchmarks, we find that complete provider traces often transfer successfully across model boundaries, but prefix trajectories reveal different mechanisms of support.
On AIME, the two receiver modes diverge sharply: in \textit{force-answer} mode, receivers usually become correct only after the prefix already contains the final answer, whereas in \textit{free-generation} mode, they often succeed before answer leakage by continuing from the partial trace.
On MMLU-Pro, transfer is more strongly shaped by receiver-intrinsic competence; on ZebraLogic, it depends on structured partial-answer information rather than complete-answer leakage alone.

Finally, we study receiver agreement as a gold-free stopping signal.
When multiple receivers give the same answer from a partial prefix, we stop and use that agreed answer.
Agreement is more reliable when the same answer persists across consecutive prefix steps, especially on AIME in \textit{force-answer} mode, where stable agreement often indicates that the prefix already contains the answer.

Our contributions are:
\begin{itemize}[leftmargin=*, noitemsep, topsep=0pt]
    \item We introduce a controlled provider--receiver study of cross-model CoT transfer across multiple models and benchmarks, using cumulative trace exposure and two trace-use modes to probe how transfer emerges.
    \item We show that transfer success does not imply a single function of the provider trace: trajectories separate answer extraction, reasoning scaffolding, receiver-intrinsic competence, and structured partial-answer information.
    \item We propose receiver agreement as a gold-free stopping signal, showing that agreement-based stopping can retain much of the transfer benefit while consuming less provider reasoning.
\end{itemize}

\section{Related Work}
\label{sec:related_work}

\begin{table*}[t]
\centering
\small
\resizebox{0.92\textwidth}{!}{
\begin{tabular}{lllrrr}
\toprule
Role & Abbr. & Model & AIME & MMLU-Pro & ZebraLogic \\
\midrule
\multirow{3}{*}{Provider}
& Qwen-4B-Thk 
& Qwen3-4B-Thinking-2507~\cite{qwen3technicalreport} 
& 84.44 & 81.22 & 96.04 \\

& Qwen-4B 
& Qwen3-4B~\cite{qwen3technicalreport} 
& 70.00 & 75.61 & 90.21 \\

& GPT 
& GPT-OSS-20B~\cite{gptoss} 
& 76.67 & 83.17 & 79.58 \\

\midrule
\multirow{4}{*}{Receiver}
& Qwen-1.7B 
& Qwen3-1.7B~\cite{qwen3technicalreport} 
& 45.56 & 63.17 & 61.88 \\

& Qwen-0.6B 
& Qwen3-0.6B~\cite{qwen3technicalreport} 
& 15.56 & 40.73 & 8.12 \\

& R1-Llama 
& DeepSeek-R1-Distill-Llama-8B~\cite{deepseekr1} 
& 36.67 & 61.95 & 24.58 \\

& SmolLM 
& SmolLM3-3B~\cite{bakouch2025smollm3} 
& 40.00 & 63.41 & 10.21 \\
\bottomrule
\end{tabular}}
\caption{Models used in the provider--receiver experiments. 
Base accuracy is reported for each model solving the benchmark with its own full reasoning trace; values are percentages.
We use subsets of those benchmarks for computational feasibility; subset selection details and statistics are described in Appendix~\ref{app:method_details}.}
\label{tab:models}
\end{table*}

\paragraph{Cross-model use of reasoning traces.}
Recent work treats reasoning traces as reusable artifacts across model boundaries:
CoTs can induce behavior in other models, be summarized or transferred across model scales, or be separated from answer execution in reuse and verification frameworks~\citep{pal2026explanationsgeneralizelargereasoning,bi2025cotxadaptiveframeworkcrossmodel,aggarwal2026evaluatingchainofthoughtreasoningreusability}.
Related collaborative and speculative reasoning systems further use intermediate reasoning steps for drafting, verification, selection, or correction across models~\citep{pan2026specreason,zhao2026trigreasontriggerbasedcollaborationsmall,shi-etal-2025-speccot,liu2026confspecefficientsteplevelspeculative}, while other work studies model behavior conditioned on provided, partial, or step-wise traces~\citep{roytburg2026measuringreasoningtracelegibility,DBLP:journals/corr/abs-2601-03154,li2026offtrajectory}.
These studies establish that reasoning traces can travel across model boundaries, but they mainly evaluate whether reuse succeeds or improves system utility.
Our work asks a different question: why and when does cross-model transfer succeed?
Using cumulative prefixes and two trace-use modes, we analyze how transfer arises from answer extraction, reasoning scaffolding, receiver-intrinsic competence, structured partial-answer information, and receiver agreement.

\paragraph{Reasoning traces as structured textual artifacts.}
Another line of work studies the internal structure of reasoning traces.
Surveys organize CoT and long-CoT methods by reasoning paradigms, structural variants, and prompting strategies~\citep{chu-etal-2024-navigate,chen2025reasoningerasurveylong}.
Other work analyzes traces to support human interpretation or oversight, for example by segmenting CoTs into supporting and opposing statements, evaluating their monitorability, identifying influential reasoning sentences, or abstracting trajectories into structured reasoning dynamics~\citep{chen-etal-2025-threading,korbak2025chainthoughtmonitorabilitynew,bogdan2025thoughtanchorsllmreasoning,yu-etal-2025-explainable}.
These studies show that CoTs have meaningful internal organization, but they largely analyze this organization from a human-facing perspective.
Our work instead uses cross-model transfer as a functional probe of trace content.
Rather than asking which parts of a CoT are interpretable to humans, we ask which partial trace information becomes useful to receiver models, and how that usefulness depends on the way receivers are allowed to use the trace.

\section{Cross-Model Reasoning Transfer}
\label{sec:cross_model_transfer}

\subsection{Benchmarks}
\label{sec:benchmarks}

We choose benchmarks to evaluate cross-model reasoning transfer.
As illustrated in Figure~\ref{fig:overview}, the three benchmarks cover complementary cases:
AIME~\cite{matharena} involves derivation-heavy mathematical reasoning with single integer answers,
MMLU-Pro~\cite{mmlupro} involves knowledge-intensive multiple-choice reasoning,
and ZebraLogic~\cite{zebralogic} involves constraint reasoning with structured multi-component answers.
These differences allow us to study whether the same provider trace can support receivers through different mechanisms, rather than treating reasoning transfer as a task-independent phenomenon.

\paragraph{Observed CoT and answer structures.}
The generated traces in Figure~\ref{fig:overview} reflect these benchmark-level differences.
AIME traces usually build toward a compact numerical answer through symbolic or quantitative derivations, so transfer may depend on when the final integer first becomes explicit.
MMLU-Pro traces often compare options using domain knowledge, so receivers may benefit from their own prior competence even when the provider prefix is short.
ZebraLogic traces gradually fill a structured solution by tracking relations among entities and attributes, so partial answer components may become useful before the full solution is available.
These differences motivate applying the provider--receiver protocol across distinct tasks.

\begin{figure*}[t]
    \centering
    \includegraphics[width=0.92\textwidth]{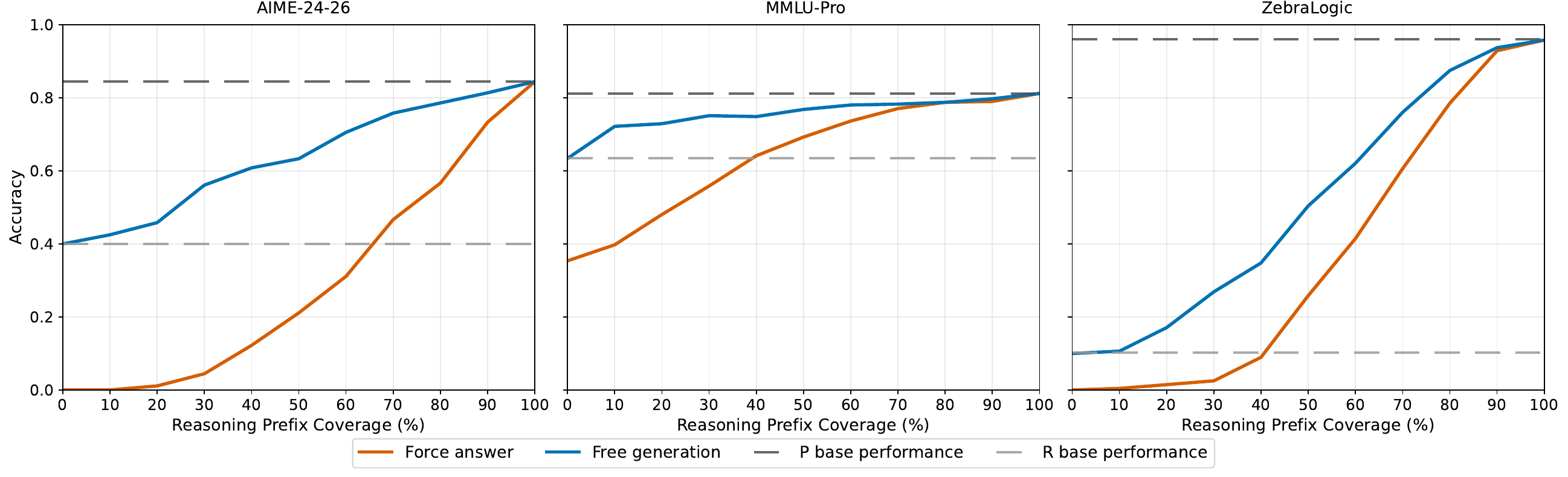}
    \caption{
Prefix-transfer trajectories for Qwen-4B as the provider and SmolLM as the receiver across datasets.
Curves show force-answer and free-generation transfer; horizontal lines mark provider and receiver base performance.
}
    \label{fig:representative_pair_3datasets}
\end{figure*}

\subsection{Models}
\label{sec:models}

We evaluate seven open-source reasoning-tuned models and assign them fixed roles as providers or receivers.
A \textit{provider} generates the reasoning trace, while a \textit{receiver} uses that trace to solve the same problem.
Our main experiments focus on higher-to-lower transfer, where stronger models provide traces to weaker receivers.
This setting is motivated by a scalable use case: if traces from stronger models can be reused by weaker models, reasoning transfer may reduce the need to run the strongest model for every answer.

Because our model pool is constrained by available compute and inference cost, our providers are not necessarily the largest available LRMs.
Instead, we assign roles by empirical performance within this pool rather than parameter count alone.
Table~\ref{tab:models} reports the models, abbreviations, roles, and base accuracy with full reasoning traces.
Provider models outperform receiver models across the benchmarks, making them suitable sources for higher-to-lower transfer in our setting.
This yields twelve cross-model transfer pairs per benchmark.

\subsection{Transfer Protocol}
\label{sec:transfer_protocol}

Full-trace transfer can show whether a provider trace is reusable, but it cannot reveal when the trace becomes useful or what information supports the receiver.
We therefore use prefix trajectories to measure how receiver behavior changes as more of a provider trace becomes available.
For each problem, the provider first generates a full reasoning trace.
We extract the verbalized CoT, split it into ten ordered segments, and construct cumulative prefixes, where the \(k\)-th prefix contains the first \(k\) segments.
For each prefix, the receiver solves the same problem under one of two trace-use modes.

In \textit{force-answer mode}, the prefix is closed as a completed reasoning segment, so the receiver must answer from the provided text.
In \textit{free-generation mode}, the prefix is left open, allowing the receiver to continue reasoning before answering.
This contrast separates what can be extracted from the available prefix from what can be scaffolded by continuing the partial trace.
This procedure is run for all provider--receiver pairs, datasets, and prefix steps.

\paragraph{Evaluation.}
As answer formats differ across benchmarks, we use benchmark-specific answer cues and extraction rules.
Predictions are evaluated by exact match for AIME, option-label match for MMLU-Pro, and full structured-solution match for ZebraLogic.
For each prefix, we aggregate multiple sampled predictions by majority vote.
Full implementation details are provided in Appendix~\ref{app:method_details}.

\begin{figure*}[t]
    \centering
    \includegraphics[width=0.92\textwidth]{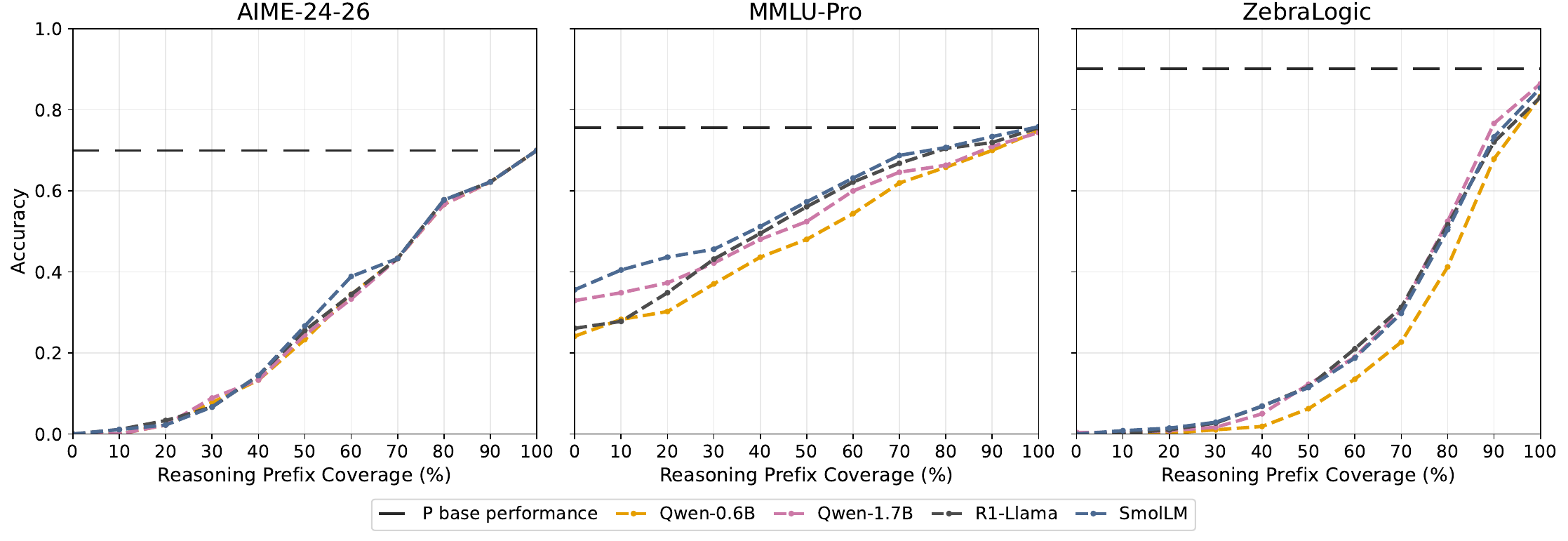}
    \caption{
Fixed-provider transfer in force-answer mode with Qwen-4B as the provider and multiple receivers across datasets.
Horizontal lines mark provider base performance.}
    \label{fig:p_multi_r_closed}
\end{figure*}

\section{Does Reasoning Transfer Successfully?}
\label{sec:transfer_exists}

We ask whether provider traces can help receivers recover provider-level performance, and how this recovery changes as more of the trace is revealed.

\paragraph{Complete traces transfer.}
When conditioned on complete provider traces, receivers approach the provider's base performance in both {force-answer} and {free-generation} modes (Figure~\ref{fig:representative_pair_3datasets}), consistent with prior evidence that CoT traces can be reused across model boundaries~\cite{pal2026explanationsgeneralizelargereasoning, bi2025cotxadaptiveframeworkcrossmodel, aggarwal2026evaluatingchainofthoughtreasoningreusability, roytburg2026measuringreasoningtracelegibility, pan2026specreason}.
Averaged over all provider--receiver pairs and both modes, the remaining absolute accuracy gap to the provider is small: 0.17 on AIME-24--26, 0.46 on MMLU-Pro, and 3.63 on ZebraLogic\footnote{
A coarse LLM-assisted error check suggests that part of the larger ZebraLogic gap comes from answer-format errors, where the final output does not satisfy the required format.
}.
Since receivers perform substantially worse without provider traces (Table~\ref{tab:models}), these results show that complete traces substantially reduce the provider--receiver performance gap.

\paragraph{Transfer accumulates across prefixes.}
Full-trace transfer shows that provider traces are reusable, but not when useful information becomes available.
We therefore evaluate cumulative trace prefixes.
As illustrated by one representative provider--receiver pair in Figure~\ref{fig:representative_pair_3datasets}, accuracy generally increases with prefix coverage in both modes, rather than remaining flat until the final segments.
Thus, partial traces already contain task-useful information, and later segments tend to further improve the receiver's ability to recover the answer.

\paragraph{Trajectory shapes depend on task and mode.}
The prefix trajectories are not uniform.
{Free-generation} often performs better at partial prefixes, suggesting that receivers can use incomplete traces as scaffolds for continued reasoning.
By contrast, {force-answer} transfer on AIME and ZebraLogic remains low until later prefixes, while MMLU-Pro shows strong {free-generation} performance even from the zero-prefix baseline.
These differences indicate that prefix usefulness is not simply a function of trace length: it also depends on the task's answer structure, the receiver's own competence, and whether the receiver must answer directly or may continue reasoning.

Figure~\ref{fig:representative_pair_3datasets} shows one provider--receiver pair for readability; Appendix~\ref{app:all_prefix_curves} reports all prefix-accuracy trajectories.
Overall, full traces transfer and partial traces provide increasing benefit, but the resulting trajectories suggest different underlying mechanisms.
We therefore analyze prefix content and receiver behavior to explain what drives these gains.

\begin{figure*}
    \centering
    \includegraphics[width=0.92\textwidth]{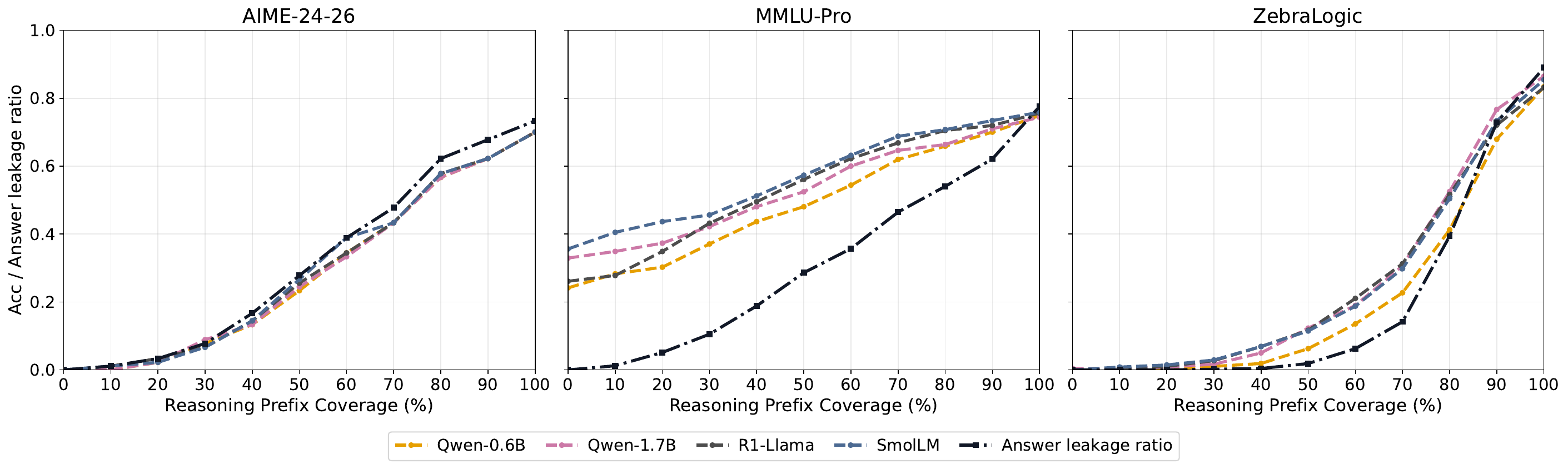}
    \caption{
Answer leakage ratio and {force-answer} transfer accuracy with Qwen-4B as the provider.
The leakage curve denotes the fraction of examples for which the correct final answer has appeared explicitly in the prefix.
}
    \label{fig:closed_leakage_curve}
\end{figure*}

\section{What Drives Force-Answer Transfer?}
\label{sec:transfer_mechanisms}

We next ask what drives the prefix-level gains observed above, starting with \textit{force-answer} mode.
Because receivers must answer from the available prefix without generating additional reasoning, this mode exposes what can be extracted from the provider trace itself.
To separate shared provider-prefix effects from receiver-specific effects, we fix the provider and compare multiple receivers under the same prefix at each coverage step.

\subsection{Force-Answer Transfer Reveals Dataset-Dependent Receiver Effects}

Figure~\ref{fig:p_multi_r_closed} shows force-answer transfer with Qwen-4B as the provider and multiple receivers across AIME, MMLU-Pro, and ZebraLogic.
Since all receivers observe the same provider prefix at each step, differences among receiver curves indicate how much transfer depends on receiver-side competence rather than only on the shared prefix.

\paragraph{AIME is dominated by shared prefix information.}
On AIME, receiver curves almost overlap throughout the prefix trajectory.
This suggests that force-answer success is mainly constrained by information already present in the provider trace, rather than by receiver-specific differences.
This pattern is consistent with AIME's multi-step mathematical structure: when continuation is disallowed, receivers cannot easily reconstruct missing reasoning and must rely on the available trace.

\paragraph{MMLU-Pro shows stronger receiver-side effects.}
MMLU-Pro exhibits a different pattern.
Receiver curves are already separated at 0\% prefix coverage and preserve a relatively stable ordering through much of the trajectory before converging near the full trace.
Thus, force-answer transfer on MMLU-Pro is not determined solely by the provider prefix; it also reflects receiver-intrinsic knowledge, option-selection behavior, and general task competence.
This is consistent with the dataset's knowledge-intensive nature, where receivers may solve or partially solve questions even with little provider reasoning~\cite{DBLP:journals/corr/abs-2601-03154}.

\paragraph{ZebraLogic depends on both prefix information and receiver use.}
ZebraLogic lies between these two cases.
Receivers start from similarly low accuracy at 0\% prefix coverage, suggesting limited task-specific ability without the provider trace.
However, their curves separate once later prefixes become informative.
Because all receivers see the same prefix, this separation suggests differences in how receivers use the available structured information, even though explicit continuation is disallowed.

These dataset-level patterns suggest that force-answer transfer can reflect different mechanisms.
AIME provides the cleanest case for isolating shared prefix information, because receiver curves closely track one another.
Inspecting first-correct prefixes shows that, in many AIME cases, the provider trace has already \textit{completed the relevant reasoning and stated the final answer.}
We therefore next test whether force-answer transfer follows explicit answer availability in the provider trace.

\subsection{Does Force-Answer Transfer Follow Answer Leakage?}

We define \textit{answer leakage} as the earliest prefix step at which the provider trace explicitly reveals the correct final answer.
For AIME, we detect leakage deterministically by checking whether the gold answer appears as a standalone numeric expression in the provider CoT.
For MMLU-Pro and ZebraLogic, we use LLM-assisted leakage detection because their answers cannot be reliably identified by exact string matching.
Details are in Appendix~\ref{app:leakage_detection}.

\paragraph{Leakage explains AIME force-answer transfer.}
Figure~\ref{fig:closed_leakage_curve} compares answer leakage with force-answer accuracy for Qwen-4B as the provider.
On AIME, the leakage curve closely tracks receiver accuracy.
Among runs where the provider is correct and the receiver becomes correct at least once, \textbf{94.5\%} of force-answer first-correct steps occur at or after answer leakage.
Thus, AIME force-answer transfer is largely explained by extracting explicit answer information from the shared provider trace.

\paragraph{Other datasets are less leakage-driven.}
MMLU-Pro and ZebraLogic show weaker alignment with complete-answer leakage.
On MMLU-Pro, 64.4\% of force-answer first-correct steps occur before the answer is explicitly available, consistent with a larger role for receiver-side knowledge and task priors.
On ZebraLogic, 40.1\% occur before the complete structured answer appears.
However, complete-answer leakage is too coarse for ZebraLogic because answers consist of multiple fields: at the component level, receivers first become correct after 82.8\% of gold answer components have appeared on average.
Thus, ZebraLogic force-answer transfer relies on dense partial-answer information rather than complete-answer leakage alone.

Overall, force-answer transfer supports a leakage-driven account for AIME, but not a single leakage-only explanation across tasks.
This raises the corresponding question for \textit{free-generation}: if receivers may continue reasoning from the same partial prefix, do they still become correct only after the answer appears?

\section{What Drives Free-Generation Transfer?}

\subsection{Free-Generation Transfer Often Succeeds Before Answer Leakage}

\begin{figure}[t]
    \centering
    \includegraphics[width=\linewidth]{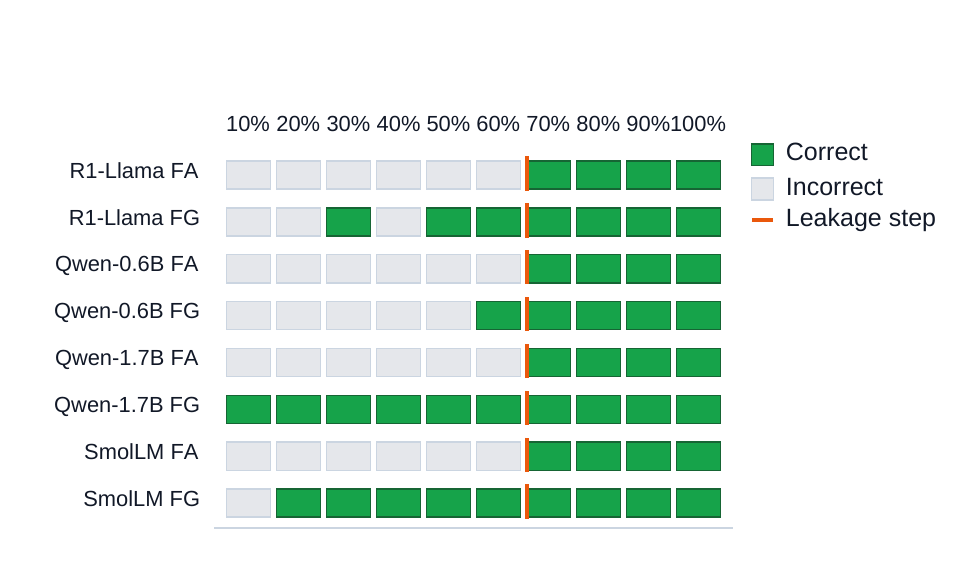}
    \caption{
AIME example comparing force-answer (FA) and free-generation (FG) transfer across steps.
Green cells indicate correct answers, gray cells indicate incorrect answers, and the red line marks the leakage step.
}
    \label{fig:leakage_case}
\end{figure}

Figure~\ref{fig:leakage_case} shows a typical AIME example around the leakage point.
The provider first states the correct answer at 70\% prefix coverage.
In force-answer mode, receivers become correct only at or after this point and often switch together.
In free-generation mode, however, several receivers answer correctly before leakage and at different prefix positions.
This mirrors that among runs where the provider is correct and the receiver becomes correct at least once, 88.11\% of free-generation first-correct steps occur before answer leakage.
Thus, most AIME free-generation successes cannot be explained by explicit answer availability alone.

For MMLU-Pro, correctness is strongly entangled with receiver-side knowledge, task competence, and option-selection behavior, making it difficult to attribute success to the prefix alone.
For ZebraLogic, component-level leakage is more informative.
Although 77.1\% of free-generation first-correct steps occur before the complete answer appears, receivers have seen only 37.7\% of gold answer components on average at those steps, compared with 82.8\% in force-answer mode.
This suggests that free-generation can use sparse structured cues as scaffolds for continued constraint reasoning, whereas force-answer transfer relies more heavily on dense partial-answer information.

These results show that free-generation transfer can occur before the answer is revealed, especially on AIME.
We therefore ask what kind of partial reasoning information helps receivers continue successfully before explicit answer leakage.

\subsection{What Information Helps Before the Answer Is Revealed?}

We focus the fine-grained prefix-content analysis on AIME, where answer leakage can be detected reliably and force-answer transfer is most closely tied to shared provider-prefix information.
To isolate useful pre-answer information, we consider cases where the receiver cannot solve the problem without a provider prefix and exclude cases where the correct answer has already appeared in the prefix.
For each remaining case, we identify the first prefix step at which the receiver becomes correct in free-generation mode.

We classify the available reasoning information into three coarse types.
\textit{Approach-level information} includes prefixes that introduce a useful representation, decomposition, equation, case split, or general line of attack.
\textit{Derivation-level information} includes concrete intermediate progress, such as derived constraints, equations, numerical relations, or partial computations.
\textit{Mixed or unclear} covers cases where both types are present or the distinction is ambiguous.

\paragraph{Both approaches and derivations help.}
We use LLM-assisted annotation and manually inspect 50 random cases, of which 42 match our judgment; we therefore treat the labels as coarse descriptive signals.
Across 494 prefixes, approach-level and derivation-level information are almost evenly split: 235 prefixes are approach-level (47.57\%), 234 are derivation-level (47.37\%), and 25 are mixed or unclear (5.06\%).
Thus, early free-generation transfer is not driven by a single type of prefix help.
Some prefixes set up a productive solution path, while others provide concrete intermediate progress.
Examples and details are provided in Appendix~\ref{app:prefix_help}.

\begin{figure}[t]
    \centering
    \includegraphics[width=\columnwidth]{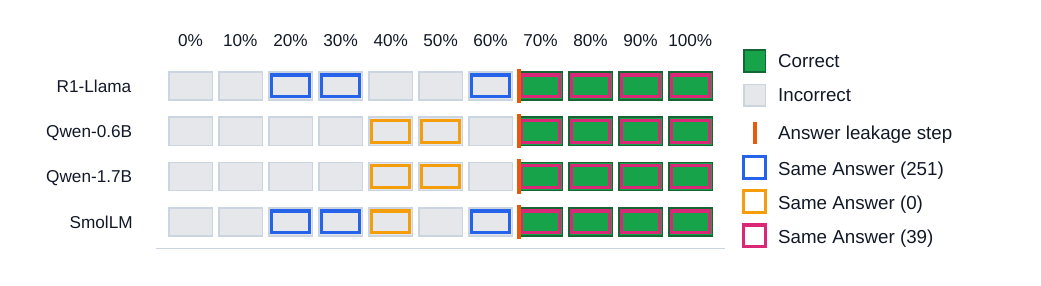}
\caption{
AIME example of receiver agreement in the force-answer mode.
Rows are receivers and columns are cumulative provider-prefix coverage.
Filled boxes show correctness, the orange vertical line marks the answer leakage step, and colored borders indicate answers shared by multiple receivers at the same prefix step.
}
    \label{fig:aime_agreement_grid}
\end{figure}

\section{Receiver Agreement as Stopping Signal}
\label{sec:agreement}

The preceding analyses show that cross-model transfer can succeed before the complete provider trace is available.
In particular, free-generation receivers often answer correctly before explicit answer leakage, while force-answer transfer on AIME becomes reliable mainly after the answer appears in the prefix.
This raises a practical question: \textit{can we detect, without access to the gold answer, when a partial provider trace is already sufficient?}

\begin{table}[t]
\centering
\scriptsize
\setlength{\tabcolsep}{4pt}
\renewcommand{\arraystretch}{0.92}
\resizebox{\columnwidth}{!}{
\begin{tabular}{@{}llcrr@{}}
\toprule
Dataset & num R & Rule & $\Delta$ Acc. Force/Free & Token Force/Free \\
\midrule
\multirow{6}{*}{AIME}
& \multirow{2}{*}{2R} & 1-step & $-26.95$ / $-9.69$ & 51.24 / 41.25 \\
&                         & 2-step & $-5.67$ / $-3.02$  & 74.01 / 64.79 \\
& \multirow{2}{*}{3R} & 1-step & $-16.45$ / $-4.72$ & 63.06 / 54.04 \\
&                         & 2-step & $-0.75$ / $-1.48$  & 79.18 / 70.89 \\
& \multirow{2}{*}{4R} & 1-step & $-11.93$ / $-3.33$ & 67.83 / 59.53 \\
&                         & 2-step & $\mathbf{+1.51}$ / $-0.74$ & 81.56 / 73.67 \\
\midrule
\multirow{6}{*}{MMLU-Pro}
& \multirow{2}{*}{2R} & 1-step & $-23.66$ / $-7.99$  & 30.45 / 57.41 \\
&                         & 2-step & $-11.37$ / $-2.97$ & 54.40 / 40.84 \\
& \multirow{2}{*}{3R} & 1-step & $-11.95$ / $-3.84$ & 45.53 / 52.95 \\
&                         & 2-step & $-4.96$ / $-1.10$  & 63.13 / 50.32 \\
& \multirow{2}{*}{4R} & 1-step & $-7.48$ / $-1.95$  & 52.04 / 53.35 \\
&                         & 2-step & $-2.76$ / $\mathbf{-0.73}$ & 67.29 / 55.86 \\
\midrule
\multirow{6}{*}{ZebraLogic}
& \multirow{2}{*}{2R} & 1-step & $-21.11$ / $-3.02$ & 77.81 / 72.79 \\
&                         & 2-step & $-2.82$ / $-0.08$  & 91.52 / 82.35 \\
& \multirow{2}{*}{3R} & 1-step & $-7.31$ / $-0.28$  & 83.66 / 78.90 \\
&                         & 2-step & $-0.97$ / $\mathbf{+0.28}$ & 93.10 / 86.96 \\
& \multirow{2}{*}{4R} & 1-step & $-3.75$ / $-0.21$  & 85.57 / 81.49 \\
&                         & 2-step & $-0.56$ / $\mathbf{+0.28}$ & 93.87 / 89.18 \\
\bottomrule
\end{tabular}}
\caption{
Accuracy--token trade-off of receiver-agreement stopping.
Each Force/Free entry reports force-answer / free-generation results.
2R--4R denote the receiver group size used for agreement.
$\Delta$ Acc. is measured relative to the provider base performance, and Token denotes the fraction of provider reasoning used before stopping.
All values are percentages.}
\label{tab:agreement_tradeoff}
\end{table}

\paragraph{Motivation.}
We consider receiver agreement as a gold-free stopping signal: if several receivers produce the same answer under the same partial provider prefix, the prefix may have constrained the answer space enough to support a reliable prediction.
Unlike correctness, agreement is observable at inference time.
We analyze it as a simulated stopping signal over fixed provider traces, rather than as a complete deployment strategy.
Figure~\ref{fig:aime_agreement_grid} shows why simple agreement is insufficient.
In this AIME example, receivers sometimes agree before leakage, but the agreed answer is unstable across prefix steps; stable four-receiver agreement appears only after the provider trace explicitly reveals the answer.
This motivates requiring agreement to persist across consecutive prefixes.

\paragraph{Stopping rules.}
At each prefix step, we collect the answers produced by a receiver group under the same provider prefix.
The \textit{one-step} rule stops at the first prefix where all receivers agree on the same answer, while the \textit{two-step} rule stops only when the same agreed answer persists for two consecutive prefix steps.
If no agreement is reached, we fall back to the complete-trace prediction.
We evaluate both rules with receiver groups of size two to four.

\paragraph{Accuracy--token trade-off.}
Table~\ref{tab:agreement_tradeoff} shows that agreement yields a task- and mode-dependent trade-off between preserving accuracy and reducing provider reasoning.
One-step agreement uses shorter prefixes but is often unreliable, whereas two-step agreement is more conservative and preserves substantially more accuracy at the cost of later stopping.
Larger receiver groups have a similar effect: they make agreement stricter and usually more reliable, but consume more provider reasoning.
Free-generation agreement often stops earlier than force-answer agreement because receivers can continue reasoning from the partial trace, although this shifts computation to the receiver side.

\paragraph{Task differences and leakage alignment.}
Agreement behavior also reflects task structure.
On MMLU-Pro, receivers can agree from shorter prefixes, consistent with the larger role of receiver-side knowledge and answer priors.
On ZebraLogic, agreement usually requires longer prefixes because receivers must align on a structured multi-component answer.
AIME explains why two-step agreement is especially useful: because force-answer success is closely tied to explicit answer availability, reliable stopping should tend to occur after leakage.
With four receivers, only 66.89\% of one-step agreements occur after leakage, compared with 91.67\% for two-step agreement; the same increase holds with three receivers (58.77\% to 89.14\%) and two receivers (42.27\% to 82.08\%).
Thus, persistent agreement filters out many unstable early agreements and better aligns stopping with informative prefixes, matching the pattern in Figure~\ref{fig:aime_agreement_grid}.
Receiver agreement turns prefix transfer analysis into a practical early-stopping heuristic.
In cost-sensitive settings where a large provider supplies expensive CoT traces to smaller receivers, stable receiver agreement can indicate when the shared prefix is already sufficient, reducing provider reasoning while retaining much of the transfer benefit.

\section{Conclusion}

We studied cross-model reasoning transfer by treating a provider trace as an unfolding sequence rather than a completed artifact. 
This prefix-based view shows that successful CoT transfer is not a single phenomenon: the same trace may serve as an explicit answer source, a scaffold for continued reasoning, or a signal whose usefulness depends on receiver competence and task structure. 
Across benchmarks, full traces often transfer well, but prefix trajectories reveal qualitatively different mechanisms, from answer extraction in force-answer AIME to pre-answer reasoning support in free generation and structured partial-answer use in ZebraLogic. 
These findings suggest that evaluating only complete-trace reuse can overstate what has been transferred and obscure how reasoning traces actually help other models. 
Finally, receiver agreement offers a practical implication of this analysis: stable agreement among smaller receivers can provide a gold-free signal for stopping provider reasoning early, pointing toward more cost-sensitive cross-model reasoning pipelines.

\section*{Limitations}

\paragraph{Prefix granularity.}
Our prefix construction uses ten cumulative segments of each provider trace. This design makes prefix trajectories comparable across examples and models, while providing one controlled way to expose partial reasoning. Different segmentation granularities, sentence boundaries, or semantic step boundaries could affect the exact prefix step at which transfer, leakage, or receiver agreement appears. We therefore interpret prefix positions as coarse indicators of when information becomes useful, rather than as precise boundaries of reasoning steps.

\paragraph{Agreement-based stopping.}
Our receiver-agreement stopping analysis measures the fraction of provider reasoning consumed. This captures how early an observable agreement signal appears along the provider trace, but end-to-end efficiency also depends on additional factors. In free-generation mode, receivers may generate further reasoning after the prefix, and practical cost depends on the relative costs of provider and receiver models, decoding length, batching, and system-level implementation details. Agreement should therefore be viewed as a useful observable signal for adaptive stopping, rather than as a complete deployment strategy.

\section*{Ethical Considerations}
This work studies chain-of-thought traces as textual artifacts that can be reused across models. We do not assume that such traces faithfully reveal a model's internal reasoning, and our findings should not be interpreted as evidence that transferred traces provide faithful explanations of model behavior. Instead, we analyze when these traces functionally help another model produce an answer.
Cross-model reasoning transfer and agreement-based stopping could reduce inference cost, but they may also make it easier to propagate incorrect or misleading reasoning from a stronger model to weaker receivers. In particular, receiver agreement is not a guarantee of correctness; multiple receivers may agree on the same wrong answer. We therefore recommend using agreement-based stopping with validation, uncertainty checks, or task-specific safeguards in high-stakes settings.

\paragraph{Reproducibility.}
Codes and outputs are publicly available at \href{https://github.com/mainlp/CoT-Transfer}{https://github.com/mainlp/CoT-Transfer} for reproduction.

\paragraph{Use of AI Assistants.}
The authors acknowledge the use of ChatGPT solely for correcting grammatical errors and enhancing the coherence of the final manuscript.

\section*{Acknowledgments}

We thank the members of the MaiNLP lab for their insightful feedback on earlier drafts of this paper. 
Beiduo Chen acknowledges his membership in the European Laboratory for Learning and Intelligent Systems (ELLIS) PhD program.
This research is supported by ERC Consolidator Grant DIALECT 101043235.


\begin{thebibliography}{33}
\providecommand{\natexlab}[1]{#1}

\bibitem[{Aggarwal et~al.(2026)Aggarwal, Mishra, and Awekar}]{aggarwal2026evaluatingchainofthoughtreasoningreusability}
Shashank Aggarwal, Ram~Vikas Mishra, and Amit Awekar. 2026.
\newblock \href {https://doi.org/10.48550/ARXIV.2602.17544} {Evaluating chain-of-thought reasoning through reusability and verifiability}.
\newblock \emph{CoRR}, abs/2602.17544.

\bibitem[{Arcuschin et~al.(2025)Arcuschin, Janiak, Krzyzanowski, Rajamanoharan, Nanda, and Conmy}]{arcuschin2025chainofthought}
Iv{\'a}n Arcuschin, Jett Janiak, Robert Krzyzanowski, Senthooran Rajamanoharan, Neel Nanda, and Arthur Conmy. 2025.
\newblock \href {https://openreview.net/forum?id=L8094Whth0} {Chain-of-thought reasoning in the wild is not always faithful}.
\newblock In \emph{Workshop on Reasoning and Planning for Large Language Models}.

\bibitem[{Bakouch et~al.(2025)Bakouch, Ben~Allal, Lozhkov, Tazi, Tunstall, Patiño, Beeching, Roucher, Reedi, Gallouédec, Rasul, Habib, Fourrier, Kydlicek, Penedo, Larcher, Morlon, Srivastav, Lochner, Nguyen, Raffel, von Werra, and Wolf}]{bakouch2025smollm3}
Elie Bakouch, Loubna Ben~Allal, Anton Lozhkov, Nouamane Tazi, Lewis Tunstall, Carlos~Miguel Patiño, Edward Beeching, Aymeric Roucher, Aksel~Joonas Reedi, Quentin Gallouédec, Kashif Rasul, Nathan Habib, Clémentine Fourrier, Hynek Kydlicek, Guilherme Penedo, Hugo Larcher, Mathieu Morlon, Vaibhav Srivastav, Joshua Lochner, and 4 others. 2025.
\newblock {SmolLM3: smol, multilingual, long-context reasoner}.
\newblock \url{https://huggingface.co/blog/smollm3}.

\bibitem[{Balunovic et~al.(2026)Balunovic, Dekoninck, Petrov, Jovanovi{\'c}, and Vechev}]{matharena}
Mislav Balunovic, Jasper Dekoninck, Ivo Petrov, Nikola Jovanovi{\'c}, and Martin Vechev. 2026.
\newblock \href {https://openreview.net/forum?id=y0zL9IZxZ7} {Matharena: Evaluating {LLM}s on uncontaminated math competitions}.
\newblock In \emph{The Thirty-ninth Annual Conference on Neural Information Processing Systems Datasets and Benchmarks Track}.

\bibitem[{Bi et~al.(2025)Bi, Chen, Wang, Hao, and Song}]{bi2025cotxadaptiveframeworkcrossmodel}
Ziqian Bi, Kaijie Chen, Tianyang Wang, Junfeng Hao, and Xinyuan Song. 2025.
\newblock \href {https://doi.org/10.48550/ARXIV.2511.05747} {Cot-x: An adaptive framework for cross-model chain-of-thought transfer and optimization}.
\newblock \emph{CoRR}, abs/2511.05747.

\bibitem[{Bogdan et~al.(2025)Bogdan, Macar, Nanda, and Conmy}]{bogdan2025thoughtanchorsllmreasoning}
Paul~C. Bogdan, Uzay Macar, Neel Nanda, and Arthur Conmy. 2025.
\newblock \href {https://doi.org/10.48550/ARXIV.2506.19143} {Thought anchors: Which {LLM} reasoning steps matter?}
\newblock \emph{CoRR}, abs/2506.19143.

\bibitem[{Chen et~al.(2026)Chen, Hu, Zhang, Litschko, Korhonen, and Plank}]{DBLP:journals/corr/abs-2601-03154}
Beiduo Chen, Tiancheng Hu, Caiqi Zhang, Robert Litschko, Anna Korhonen, and Barbara Plank. 2026.
\newblock \href {https://doi.org/10.48550/ARXIV.2601.03154} {Decoupling the effect of chain-of-thought reasoning: {A} human label variation perspective}.
\newblock \emph{CoRR}, abs/2601.03154.

\bibitem[{Chen et~al.(2025{\natexlab{a}})Chen, Liu, Korhonen, and Plank}]{chen-etal-2025-threading}
Beiduo Chen, Yang~Janet Liu, Anna Korhonen, and Barbara Plank. 2025{\natexlab{a}}.
\newblock \href {https://doi.org/10.18653/v1/2025.emnlp-main.1682} {Threading the needle: Reweaving chain-of-thought reasoning to explain human label variation}.
\newblock In \emph{Proceedings of the 2025 Conference on Empirical Methods in Natural Language Processing}, pages 33111--33135, Suzhou, China. Association for Computational Linguistics.

\bibitem[{Chen et~al.(2025{\natexlab{b}})Chen, Qin, Liu, Peng, Guan, Wang, Hu, Zhou, Gao, and Che}]{chen2025reasoningerasurveylong}
Qiguang Chen, Libo Qin, Jinhao Liu, Dengyun Peng, Jiannan Guan, Peng Wang, Mengkang Hu, Yuhang Zhou, Te~Gao, and Wanxiang Che. 2025{\natexlab{b}}.
\newblock \href {https://doi.org/10.48550/ARXIV.2503.09567} {Towards reasoning era: {A} survey of long chain-of-thought for reasoning large language models}.
\newblock \emph{CoRR}, abs/2503.09567.

\bibitem[{Chen et~al.(2025{\natexlab{c}})Chen, Benton, Radhakrishnan, Uesato, Denison, Schulman, Somani, Hase, Wagner, Roger, Mikulik, Bowman, Leike, Kaplan, and Perez}]{chen2025reasoning}
Yanda Chen, Joe Benton, Ansh Radhakrishnan, Jonathan Uesato, Carson Denison, John Schulman, Arushi Somani, Peter Hase, Misha Wagner, Fabien Roger, Vladimir Mikulik, Samuel~R. Bowman, Jan Leike, Jared Kaplan, and Ethan Perez. 2025{\natexlab{c}}.
\newblock \href {https://doi.org/10.48550/ARXIV.2505.05410} {Reasoning models don't always say what they think}.
\newblock \emph{CoRR}, abs/2505.05410.

\bibitem[{Chu et~al.(2024)Chu, Chen, Chen, Yu, He, Wang, Peng, Liu, Qin, and Liu}]{chu-etal-2024-navigate}
Zheng Chu, Jingchang Chen, Qianglong Chen, Weijiang Yu, Tao He, Haotian Wang, Weihua Peng, Ming Liu, Bing Qin, and Ting Liu. 2024.
\newblock \href {https://doi.org/10.18653/v1/2024.acl-long.65} {Navigate through enigmatic labyrinth a survey of chain of thought reasoning: Advances, frontiers and future}.
\newblock In \emph{Proceedings of the 62nd Annual Meeting of the Association for Computational Linguistics (Volume 1: Long Papers)}, pages 1173--1203, Bangkok, Thailand. Association for Computational Linguistics.

\bibitem[{DeepSeek-AI(2026)}]{deepseekapi}
DeepSeek-AI. 2026.
\newblock Deepseek-v4: Towards highly efficient million-token context intelligence.

\bibitem[{Ettinger et~al.(2025)Ettinger, Bertsch, Kuehl, Graham, Heineman, Groeneveld, Brahman, Timbers, Ivison, Morrison, Poznanski, Lo, Soldaini, Jordan, Chen, Noukhovitch, Lambert, Walsh, Dasigi, Berry, Malik, Shah, Geng, Arora, Gupta, Anderson, Xiao, Murray, Romero, Graf, Asai, Bhagia, Wettig, Liu, Rangapur, Anastasiades, Huang, Schwenk, Trivedi, Magnusson, Lochner, Liu, Miranda, Sap, Morgan, Schmitz, Guerquin, Wilson, Huff, Bras, Xin, Shao, Skjonsberg, Shen, Li, Wilde, Pyatkin, Merrill, Chang, Gu, Zeng, Sabharwal, Zettlemoyer, Koh, Farhadi, Smith, and Hajishirzi}]{olmo2026olmo3}
Allyson Ettinger, Amanda Bertsch, Bailey Kuehl, David Graham, David Heineman, Dirk Groeneveld, Faeze Brahman, Finbarr Timbers, Hamish Ivison, Jacob Morrison, Jake Poznanski, Kyle Lo, Luca Soldaini, Matt Jordan, Mayee~F. Chen, Michael Noukhovitch, Nathan Lambert, Pete Walsh, Pradeep Dasigi, and 48 others. 2025.
\newblock \href {https://doi.org/10.48550/ARXIV.2512.13961} {Olmo 3}.
\newblock \emph{CoRR}, abs/2512.13961.

\bibitem[{Guo et~al.(2025)Guo, Yang, Zhang, Song, Wang, Zhu, Xu, Zhang, Ma, Bi, Zhang, Yu, Wu, Wu, Gou, Shao, Li, Gao, Liu, Xue, Wang, Wu, Feng, Lu, Zhao, Deng, Ruan, Dai, Chen, Ji, Li, Lin, Dai, Luo, Hao, Chen, Li, Zhang, Xu, Ding, Gao, Qu, Li, Guo, Li, Chen, Yuan, Tu, Qiu, Li, Cai, Ni, Liang, Chen, Dong, Hu, You, Gao, Guan, Huang, Yu, Wang, Zhang, Zhao, Wang, Zhang, Xu, Xia, Zhang, Zhang, Tang, Zhou, Li, Wang, Li, Tian, Huang, Zhang, Wang, Chen, Du, Ge, Zhang, Pan, Wang, Chen, Jin, Chen, Lu, Zhou, Chen, Ye, Wang, Yu, Zhou, Pan, Li, Zhou, Wu, Yun, Pei, Sun, Wang, Zeng, Liu, Liang, Gao, Yu, Zhang, Xiao, An, Liu, Wang, Chen, Nie, Cheng, Liu, Xie, Liu, Yang, Li, Su, Lin, Li, Jin, Shen, Chen, Sun, Wang, Song, Zhou, Wang, Shan, Li, Wang, Wei, Zhang, Xu, Li, Zhao, Sun, Wang, Yu, Zhang, Shi, Xiong, He, Piao, Wang, Tan, Ma, Liu, Guo, Ou, Wang, Gong, Zou, He, Xiong, Luo, You, Liu, Zhou, Zhu, Huang, Li, Zheng, Zhu, Ma, Tang, Zha, Yan, Ren, Ren, Sha, Fu, Xu, Xie, Zhang, Hao, Ma, Yan, Wu, Gu, Zhu, Liu, Li, Xie, Song,
  Pan, Huang, Xu, Zhang, and Zhang}]{deepseekr1}
Daya Guo, Dejian Yang, Haowei Zhang, Junxiao Song, Peiyi Wang, Qihao Zhu, Runxin Xu, Ruoyu Zhang, Shirong Ma, Xiao Bi, Xiaokang Zhang, Xingkai Yu, Yu~Wu, Z.~F. Wu, Zhibin Gou, Zhihong Shao, Zhuoshu Li, Ziyi Gao, Aixin Liu, and 175 others. 2025.
\newblock \href {https://doi.org/10.1038/S41586-025-09422-Z} {Deepseek-r1 incentivizes reasoning in llms through reinforcement learning}.
\newblock \emph{Nat.}, 645(8081):633--638.

\bibitem[{Korbak et~al.(2025)Korbak, Balesni, Barnes, Bengio, Benton, Bloom, Chen, Cooney, Dafoe, Dragan, Emmons, Evans, Farhi, Greenblatt, Hendrycks, Hobbhahn, Hubinger, Irving, Jenner, Kokotajlo, Krakovna, Legg, Lindner, Luan, Madry, Michael, Nanda, Orr, Pachocki, Perez, Phuong, Roger, Saxe, Shlegeris, Soto, Steinberger, Wang, Zaremba, Baker, Shah, and Mikulik}]{korbak2025chainthoughtmonitorabilitynew}
Tomek Korbak, Mikita Balesni, Elizabeth Barnes, Yoshua Bengio, Joe Benton, Joseph Bloom, Mark Chen, Alan Cooney, Allan Dafoe, Anca~D. Dragan, Scott Emmons, Owain Evans, David Farhi, Ryan Greenblatt, Dan Hendrycks, Marius Hobbhahn, Evan Hubinger, Geoffrey Irving, Erik Jenner, and 22 others. 2025.
\newblock \href {https://doi.org/10.48550/ARXIV.2507.11473} {Chain of thought monitorability: {A} new and fragile opportunity for {AI} safety}.
\newblock \emph{CoRR}, abs/2507.11473.

\bibitem[{Lanham et~al.(2023)Lanham, Chen, Radhakrishnan, Steiner, Denison, Hernandez, Li, Durmus, Hubinger, Kernion, Lukosiute, Nguyen, Cheng, Joseph, Schiefer, Rausch, Larson, McCandlish, Kundu, Kadavath, Yang, Henighan, Maxwell, Telleen{-}Lawton, Hume, Hatfield{-}Dodds, Kaplan, Brauner, Bowman, and Perez}]{lanham2023measuringfaithfulnesschainofthoughtreasoning}
Tamera Lanham, Anna Chen, Ansh Radhakrishnan, Benoit Steiner, Carson Denison, Danny Hernandez, Dustin Li, Esin Durmus, Evan Hubinger, Jackson Kernion, Kamile Lukosiute, Karina Nguyen, Newton Cheng, Nicholas Joseph, Nicholas Schiefer, Oliver Rausch, Robin Larson, Sam McCandlish, Sandipan Kundu, and 11 others. 2023.
\newblock \href {https://doi.org/10.48550/ARXIV.2307.13702} {Measuring faithfulness in chain-of-thought reasoning}.
\newblock \emph{CoRR}, abs/2307.13702.

\bibitem[{Li and Goyal(2026)}]{li2026offtrajectory}
Aochong~Oliver Li and Tanya Goyal. 2026.
\newblock \href {https://openreview.net/forum?id=hVUIguIm14} {Off-trajectory reasoning: Can {LLM}s collaborate on reasoning trajectories?}
\newblock In \emph{The Fourteenth International Conference on Learning Representations}.

\bibitem[{Lin et~al.(2025)Lin, Bras, Richardson, Sabharwal, Poovendran, Clark, and Choi}]{zebralogic}
Bill~Yuchen Lin, Ronan~Le Bras, Kyle Richardson, Ashish Sabharwal, Radha Poovendran, Peter Clark, and Yejin Choi. 2025.
\newblock \href {https://proceedings.mlr.press/v267/lin25i.html} {Zebralogic: On the scaling limits of llms for logical reasoning}.
\newblock In \emph{Forty-second International Conference on Machine Learning, {ICML} 2025, Vancouver, BC, Canada, July 13-19, 2025}, Proceedings of Machine Learning Research. {PMLR} / OpenReview.net.

\bibitem[{Liu and He(2026)}]{liu2026confspecefficientsteplevelspeculative}
Siran Liu and Cyril~Y. He. 2026.
\newblock \href {https://doi.org/10.48550/ARXIV.2602.18447} {Confspec: Efficient step-level speculative reasoning via confidence-gated verification}.
\newblock \emph{CoRR}, abs/2602.18447.

\bibitem[{Mondorf and Plank(2024{\natexlab{a}})}]{mondorf2024beyond}
Philipp Mondorf and Barbara Plank. 2024{\natexlab{a}}.
\newblock \href {https://openreview.net/forum?id=Lmjgl2n11u} {Beyond accuracy: Evaluating the reasoning behavior of large language models - a survey}.
\newblock In \emph{First Conference on Language Modeling}.

\bibitem[{Mondorf and Plank(2024{\natexlab{b}})}]{mondorf-plank-2024-comparing}
Philipp Mondorf and Barbara Plank. 2024{\natexlab{b}}.
\newblock \href {https://doi.org/10.18653/v1/2024.acl-long.508} {Comparing inferential strategies of humans and large language models in deductive reasoning}.
\newblock In \emph{Proceedings of the 62nd Annual Meeting of the Association for Computational Linguistics (Volume 1: Long Papers)}, pages 9370--9402, Bangkok, Thailand. Association for Computational Linguistics.

\bibitem[{OpenAI(2024)}]{gpto1}
OpenAI. 2024.
\newblock \href {https://doi.org/10.48550/ARXIV.2412.16720} {Openai o1 system card}.
\newblock \emph{CoRR}, abs/2412.16720.

\bibitem[{OpenAI(2025)}]{gptoss}
OpenAI. 2025.
\newblock \href {https://arxiv.org/abs/2508.10925} {gpt-oss-120b \& gpt-oss-20b model card}.
\newblock \emph{Preprint}, arXiv:2508.10925.

\bibitem[{Pal et~al.(2026)Pal, Bau, and Singh}]{pal2026explanationsgeneralizelargereasoning}
Koyena Pal, David Bau, and Chandan Singh. 2026.
\newblock \href {https://doi.org/10.48550/ARXIV.2601.11517} {Do explanations generalize across large reasoning models?}
\newblock \emph{CoRR}, abs/2601.11517.

\bibitem[{Pan et~al.(2026)Pan, Dai, Zhang, Oliaro, Jia, and Netravali}]{pan2026specreason}
Rui Pan, Yinwei Dai, Zhihao Zhang, Gabriele Oliaro, Zhihao Jia, and Ravi Netravali. 2026.
\newblock \href {https://openreview.net/forum?id=wCbOKbZ7kf} {Specreason: Fast and accurate inference-time compute via speculative reasoning}.
\newblock In \emph{The Thirty-ninth Annual Conference on Neural Information Processing Systems}.

\bibitem[{Paul et~al.(2024)Paul, West, Bosselut, and Faltings}]{paul-etal-2024-making}
Debjit Paul, Robert West, Antoine Bosselut, and Boi Faltings. 2024.
\newblock \href {https://doi.org/10.18653/v1/2024.findings-emnlp.882} {Making reasoning matter: Measuring and improving faithfulness of chain-of-thought reasoning}.
\newblock In \emph{Findings of the Association for Computational Linguistics: EMNLP 2024}, pages 15012--15032, Miami, Florida, USA. Association for Computational Linguistics.

\bibitem[{Roytburg et~al.(2026)Roytburg, Sridhar, and Ippolito}]{roytburg2026measuringreasoningtracelegibility}
Dani Roytburg, Shreya Sridhar, and Daphne Ippolito. 2026.
\newblock \href {https://doi.org/10.48550/ARXIV.2603.20508} {Measuring reasoning trace legibility: Can those who understand teach?}
\newblock \emph{CoRR}, abs/2603.20508.

\bibitem[{Shi et~al.(2025)Shi, Zhu, Shi, Zhao, Li, and Jiang}]{shi-etal-2025-speccot}
Junhan Shi, Yijia Zhu, Zhenning Shi, Dan Zhao, Qing Li, and Yong Jiang. 2025.
\newblock \href {https://doi.org/10.18653/v1/2025.findings-emnlp.1326} {{S}pec{C}o{T}: Accelerating chain-of-thought reasoning through speculative exploration}.
\newblock In \emph{Findings of the Association for Computational Linguistics: EMNLP 2025}, pages 24405--24415, Suzhou, China. Association for Computational Linguistics.

\bibitem[{Team(2025)}]{qwen3technicalreport}
Qwen Team. 2025.
\newblock \href {https://doi.org/10.48550/ARXIV.2505.09388} {Qwen3 technical report}.
\newblock \emph{CoRR}, abs/2505.09388.

\bibitem[{Turpin et~al.(2023)Turpin, Michael, Perez, and Bowman}]{turpin2023faithful}
Miles Turpin, Julian Michael, Ethan Perez, and Samuel~R. Bowman. 2023.
\newblock \href {http://papers.nips.cc/paper\_files/paper/2023/hash/ed3fea9033a80fea1376299fa7863f4a-Abstract-Conference.html} {Language models don't always say what they think: Unfaithful explanations in chain-of-thought prompting}.
\newblock In \emph{Advances in Neural Information Processing Systems 36: Annual Conference on Neural Information Processing Systems 2023, NeurIPS 2023, New Orleans, LA, USA, December 10 - 16, 2023}.

\bibitem[{Wang et~al.(2024)Wang, Ma, Zhang, Ni, Chandra, Guo, Ren, Arulraj, He, Jiang, Li, Ku, Wang, Zhuang, Fan, Yue, and Chen}]{mmlupro}
Yubo Wang, Xueguang Ma, Ge~Zhang, Yuansheng Ni, Abhranil Chandra, Shiguang Guo, Weiming Ren, Aaran Arulraj, Xuan He, Ziyan Jiang, Tianle Li, Max Ku, Kai Wang, Alex Zhuang, Rongqi Fan, Xiang Yue, and Wenhu Chen. 2024.
\newblock \href {http://papers.nips.cc/paper\_files/paper/2024/hash/ad236edc564f3e3156e1b2feafb99a24-Abstract-Datasets\_and\_Benchmarks\_Track.html} {Mmlu-pro: {A} more robust and challenging multi-task language understanding benchmark}.
\newblock In \emph{Advances in Neural Information Processing Systems 38: Annual Conference on Neural Information Processing Systems 2024, NeurIPS 2024, Vancouver, BC, Canada, December 10 - 15, 2024}.

\bibitem[{Yu et~al.(2025)Yu, Xiong, Wu, Li, Yu, Chen, Sinha, Shang, and McAuley}]{yu-etal-2025-explainable}
Sheldon Yu, Yuxin Xiong, Junda Wu, Xintong Li, Tong Yu, Xiang Chen, Ritwik Sinha, Jingbo Shang, and Julian McAuley. 2025.
\newblock \href {https://doi.org/10.18653/v1/2025.findings-emnlp.904} {Explainable chain-of-thought reasoning: An empirical analysis on state-aware reasoning dynamics}.
\newblock In \emph{Findings of the Association for Computational Linguistics: EMNLP 2025}, pages 16660--16667, Suzhou, China. Association for Computational Linguistics.

\bibitem[{Zhao et~al.(2026)Zhao, Peng, Nguyen, Li, Wang, Fu, and Zhao}]{zhao2026trigreasontriggerbasedcollaborationsmall}
Yi~Zhao, Yajuan Peng, Cam{-}Tu Nguyen, Zuchao Li, Xiaoliang Wang, Xiaoming Fu, and Hai Zhao. 2026.
\newblock \href {https://doi.org/10.48550/ARXIV.2604.14847} {Trigreason: Trigger-based collaboration between small and large reasoning models}.
\newblock \emph{CoRR}, abs/2604.14847.

\end{thebibliography}

\appendix

\section{Additional Implementation Details}
\label{app:method_details}

\subsection{Prompts and Answer Elicitation}

\paragraph{Base prompts.}
We use dataset-specific base prompts before applying the model-specific chat template.
These prompts are used to generate the provider reasoning traces.

For AIME-24--26, the base prompt asks the model to reason step by step and put the final integer answer in:

\begin{greenpromptbox}{AIME base prompt}
Please reason step by step, and put your final answer within \textbackslash boxed\{\}.\\
The answer is an integer between 0 and 999 inclusive.\\
\{problem\}
\end{greenpromptbox}

For MMLU-Pro, the base prompt presents a multiple-choice question with answer options A--J and asks the model to finish with \texttt{the answer is (X)}:

\begin{greenpromptbox}{MMLU-Pro base prompt}
The following are multiple choice questions (with answers) about \{subject\}.\\
Think step by step and then finish your answer with ``the answer is (X)''\\
where X is the correct letter choice.\\
Question: \{question\}\\
Options: A. \{option\_1\} B. \{option\_2\} ... J. \{option\_10\}\\
Answer:
\end{greenpromptbox}

For ZebraLogic, the base prompt follows the original format: one solved example puzzle and its JSON solution are followed by the target puzzle and the required JSON schema:

\begin{greenpromptbox}{ZebraLogic base prompt}
\# Example Puzzle\\
{}[fixed solved example]\\
\#\# Answer to the Example Puzzle\\
{}[fixed JSON solution]\\
\# Puzzle to Solve\\
\{PUZZLE\}\\
\# Instruction\\
Now please solve the above puzzle. Present only the solution in the following json format:\\
\{SCHEMA\}
\end{greenpromptbox}

Here, \texttt{\{problem\}}, \texttt{\{question\}}, and \texttt{\{PUZZLE\}} are replaced by the corresponding benchmark instance.
For ZebraLogic, \texttt{\{SCHEMA\}} specifies the required JSON structure for the current puzzle, with placeholder values to be filled by the model.

Although the Figure~\ref{fig:overview} uses \texttt{<think>} and \texttt{</think>} as notation for reasoning boundaries, these markers are not intended to impose a single literal format across models.
In implementation, each model is prompted using its native chat template and reasoning format.

\paragraph{Dataset language.}
All benchmarks are English-language or English-formatted: AIME-24--26 uses English mathematical word problems, MMLU-Pro uses English multiple-choice questions, and ZebraLogic uses English constraint clues with JSON-formatted answers. 

\paragraph{Answer elicitation cues.}
After the reasoning context is closed, we elicit the final answer using benchmark-specific cues.
The same cues are used in the force-answer mode and after the generated continuation in the free-generation mode.
Table~\ref{tab:answer_cues} summarizes the cues and expected answer formats.

\begin{table}[t]
\centering
\scriptsize
\setlength{\tabcolsep}{3pt}
\renewcommand{\arraystretch}{0.92}
\begin{tabularx}{\columnwidth}{@{}lXl@{}}
\toprule
Benchmark & Answer cue & Expected format \\
\midrule
AIME
& \texttt{Based on the reasoning so far, the final answer is: \textbackslash boxed\{}
& \texttt{\textbackslash boxed\{integer\}} \\

MMLU-Pro
& \texttt{Based on the reasoning so far, the final answer is (}
& \texttt{(X)} label \\

ZebraLogic
& \texttt{Based on the reasoning so far, the final solution JSON format is}
& JSON object \\
\bottomrule
\end{tabularx}
\caption{Benchmark-specific answer cues used after the provided or generated reasoning context.}
\label{tab:answer_cues}
\end{table}

\subsection{Prefix Construction and Decoding}
\label{app:prefix_decoding}

\paragraph{Prefix segmentation.}
We extract only the verbalized reasoning content between each model’s reasoning markers and segment this extracted trace into ten ordered segments. Sentence boundaries are obtained using NLTK sentence tokenization. Token lengths are computed with the provider model’s tokenizer. We then use dynamic programming to partition the sentence sequence into ten consecutive segments by minimizing the sum of squared deviations between each segment’s token length and the target segment length, where the target is the total number of provider-tokenized reasoning tokens divided by ten. This procedure encourages balanced token coverage while preserving sentence boundaries. If sentence-level segmentation cannot produce ten valid segments, for example because the extracted reasoning trace contains fewer than ten sentence units, we fall back to token-level partitioning and split the provider-tokenized reasoning trace into ten approximately equal consecutive spans. The ten segments are then concatenated cumulatively to form the prefix sequence used in receiver evaluation.

\paragraph{Two-stage evaluation.}
All experiments follow a two-stage procedure.
First, each provider model is run once on the original dataset prompt to generate a complete reasoning trace for each problem.
Second, the extracted trace is segmented into cumulative prefixes and used for prefix-conditioned receiver evaluation.
This ensures that receiver behavior is evaluated against fixed provider traces rather than traces regenerated separately for each prefix condition.

\paragraph{Decoding and aggregation.}
In the force-answer mode, each prefix condition is decoded five times for final-answer elicitation.
In the free-generation mode, each prefix condition is decoded five times to generate five reasoning continuations, and the final answer is elicited once from each continuation using the benchmark-specific answer cue.
We use a single answer-elicitation pass in free-generation mode because a pilot check on AIME showed that repeated answer elicitation from the same generated continuation was highly stable, with 96.44\% of conditions yielding unanimous answers across five repeated elicitation runs.

In both modes, extracted answers are aggregated by majority vote.
Ties are resolved by fixed-seed random tie-breaking.

\subsection{Answer Extraction, Generation Parameters, and Compute}
\label{app:extraction_compute}

\paragraph{Answer extraction.}
We use benchmark-specific answer extraction rules that match the expected answer format induced by the corresponding answer cue.
For AIME-24--26, we extract the integer inside the final \texttt{\textbackslash boxed\{\}} expression.
For MMLU-Pro, we extract the option label following the opening parenthesis introduced by the answer cue.
For ZebraLogic, we complete the JSON prefix introduced by the cue, parse the resulting JSON object, and compare the structured assignment against the gold solution.
A ZebraLogic prediction is counted as correct only if all required fields are parsed and match the gold assignment.

\paragraph{Generation parameters.}
Table~\ref{tab:generation_params} summarizes the default generation parameters used in the main experiments.
We use the default sampling configuration specified for each model family, following the developers' recommended settings where available.
This is intended to preserve each model's native reasoning behavior and Chain-of-Thought (CoT) generation style.
Unless otherwise stated, all experiments use a maximum generation length of 32,768 tokens.

\begin{table}[t]
\centering
\small
\resizebox{\linewidth}{!}{
\begin{tabular}{lccccc}
\toprule
Model & Temp. & Top-$p$ & Top-$k$ & Min-$p$ & Max tokens \\
\midrule
Qwen-4B-Thk & 0.6 & 0.95 & 20 & 0 & 32768 \\
Qwen-4B & 0.6 & 0.95 & 20 & 0 & 32768 \\
GPT & 0.6 & 0.95 & -- & -- & 32768 \\
Qwen-1.7B & 0.6 & 0.95 & 20 & 0 & 32768 \\
Qwen-0.6B & 0.6 & 0.95 & 20 & 0 & 32768 \\
R1-Llama & 0.6 & 0.95 & -- & -- & 32768 \\
SmolLM & 0.6 & 0.95 & -- & -- & 32768 \\
\bottomrule
\end{tabular}
}
\caption{Default generation parameters used in the main experiments. Model abbreviations follow Table~\ref{tab:models}.}
\label{tab:generation_params}
\end{table}

\paragraph{Compute resources.}
All experiments were conducted on four NVIDIA A100-SXM4-80GB GPUs.
The complete set of main experiments took approximately 120 hours of wall-clock time on this setup.

\begin{table}[t]
\centering
\small
\begin{tabular}{lcl}
\toprule
Benchmark & \#Items & Evaluation format \\
\midrule
AIME-24--26 & 90 & Integer exact match \\
ZebraLogic & 480 & Full structured JSON match \\
MMLU-Pro CS & 410 & 10-option multiple choice \\
\bottomrule
\end{tabular}
\caption{Benchmark subsets used in the experiments. All datasets are in English.}
\label{tab:benchmark_subsets}
\end{table}

\section{Answer Leakage Detection}
\label{app:leakage_detection}

We define answer leakage as the earliest prefix segment at which the provider trace explicitly reveals the correct final answer.
For AIME, leakage is detected deterministically by checking whether the gold integer answer appears as a standalone numeric expression in the provider reasoning text.
For MMLU-Pro and ZebraLogic, we use LLM-assisted leakage detection because the correct answer may be expressed as an option, a paraphrased statement, or a structured solution rather than a single exact string.

We probe prefix leakage by asking a separate DeepSeek judge model (\texttt{deepseek-v4-flash})~\cite{deepseekapi} to identify the earliest segment in which the complete gold answer is explicitly stated.
The judge is run with temperature 0, JSON-only output, and thinking disabled.
For MMLU-Pro and ZebraLogic, the judge is instructed to mark leakage only when the complete correct final answer is explicitly stated in one continuous answer statement.
It must not solve the task, infer unstated answers, or combine partial information across separated reasoning steps.
If an answer statement is split across adjacent segments because the segmentation boundary cuts through an uninterrupted answer span, the leakage point is assigned to the later segment where the full answer becomes available.

For ZebraLogic, we additionally run a partial-leakage judge that identifies which gold answer cells become explicitly correct for the first time in each segment.
A cell is counted only when the full house--attribute--value assignment is explicitly stated.
The judge returns newly explicit cells per segment rather than cumulative cells, and the same cell is not counted more than once.

To assess reliability, we manually inspected 50 examples from each dataset using LLM-assisted leakage detection.
As shown in Table~\ref{tab:leakage_validation}, the LLM-assisted labels agreed with the manual inspection in all checked cases.
We use this validation as a reliability check rather than as a source of final labels.

\begin{table}[t]
\centering
\small
\begin{tabular}{lcc}
\toprule
Dataset & \#Checked & Agreement \\
\midrule
MMLU-Pro & 50 & 100\% \\
ZebraLogic & 50 & 100\% \\
\bottomrule
\end{tabular}
\caption{Manual validation of LLM-assisted answer-leakage detection.}
\label{tab:leakage_validation}
\end{table}

\paragraph{Full-answer leakage judge prompt.}
The following prompt is used for MMLU-Pro and ZebraLogic full-answer leakage detection.

\textbf{System prompt.}
\begin{bluepromptbox}{System prompt}
You are a strict annotation judge.\\
Your job is only to judge whether the reasoning explicitly states the complete correct answer in one continuous answer statement.\\
You may treat adjacent segments as one continuous statement only when the segmentation cut through an uninterrupted answer span.\\
When a continuous answer statement crosses from one segment into the next, assign the hit to the later segment where the full answer becomes complete.\\
Do not mark the earlier segment positive if it contains only an incomplete prefix.\\
Do not combine partial information across separated reasoning steps.\\
Do not solve the task yourself.\\
Do not infer the answer from the problem if the text itself has not already made it explicit.\\
Return JSON only.
\end{bluepromptbox}

\textbf{User prompt template.}
\begin{orangepromptbox}{User prompt template}
Below is one model's reasoning for a single problem, split into 10 segments in chronological order.\\

Your task is to identify the earliest segment from which the reasoning already explicitly states the complete correct final answer.\\

Rules:\\
1. Judge only what is explicitly stated in the reasoning text.\\
2. Do not solve the problem yourself.\\
3. Do not give credit for hints, partial narrowing, or suggestive evidence.\\
4. Count it as positive if the complete correct answer is explicitly stated in one continuous answer statement.\\
5. A continuous answer statement may cross an adjacent segment boundary if the segmentation merely cut one uninterrupted sentence or answer span into pieces.\\
6. If a continuous answer statement crosses from segment N into segment N+1, count the leakage at segment N+1, because that is the segment where the full answer first becomes available.\\
7. Do not mark segment N positive if it only contains an incomplete prefix of the answer and the full answer is completed only in segment N+1.\\
8. Do not combine information across separated reasoning steps. If one segment mentions part of the answer, then there is other reasoning, and only later another segment adds the rest, that does not count.\\
9. If the answer is never made explicit in one continuous answer statement, return null.\\

Correct final answer:\\
\{GOLD\_ANSWER\}\\

Reasoning segments:\\
{}[Segment 1]\\
\{SEGMENT\_1\}\\

{}[Segment 2]\\
\{SEGMENT\_2\}\\

...\\

{}[Segment 10]\\
\{SEGMENT\_10\}\\

Return JSON only with this schema:\\
\{\\
\ \ "earliest\_explicit\_segment": 1-10 or null,\\
\ \ "segment\_assessments": [\\
\ \ \ \ \{\\
\ \ \ \ \ \ "segment\_idx": 1,\\
\ \ \ \ \ \ "explicit\_correct\_answer": true/false,\\
\ \ \ \ \ \ "reason": "short explanation"\\
\ \ \ \ \}\\
\ \ ]\\
\}
\end{orangepromptbox}

\paragraph{ZebraLogic partial-leakage judge prompt.}
For ZebraLogic, we additionally identify partial leakage at the level of gold answer cells.
The judge receives the gold cell IDs and the ten reasoning segments, and returns the newly explicit correct cell IDs for each segment.

\textbf{System prompt.}
\begin{bluepromptbox}{System prompt}
You are a strict annotation judge.\\
Your job is only to judge which gold answer cells become explicitly correct for the first time in each segment.\\
Do not solve the puzzle yourself.\\
Do not infer unstated cells.\\
Only count cells whose full house-attribute-value assignment is explicitly stated.\\
Output incremental per-segment newly explicit cells, not cumulative cells.\\
Do not count the same cell again after it first appears.\\
If an uninterrupted answer statement is cut across adjacent segments, assign the completed cell(s) to the later segment where the full statement becomes available.\\
Use only the provided gold cell\_ids.\\
Return JSON only.
\end{bluepromptbox}

\textbf{User prompt template.}
\begin{orangepromptbox}{User prompt template}
Below is one model's reasoning for a single ZebraLogic problem, split into 10 chronological segments.\\

Your task is to judge incremental partial answer leakage for a ZebraLogic solution.\\
A gold answer cell is one exact house-attribute-value assignment.\\
For each segment, identify which gold answer cells become explicitly correct for the first time in that segment.\\

Rules:\\
1. Judge only what is explicitly stated in the reasoning text.\\
2. Do not solve the puzzle yourself.\\
3. Count a cell only if the full house-attribute-value assignment is explicitly stated.\\
4. Do not infer cells from hints, eliminations, or other stated cells.\\
5. If one cell's full statement is split across two adjacent segments, count it in the later segment where it becomes complete.\\
6. Return exactly 10 items: one for each segment 1 through 10.\\
7. If a segment has no newly explicit correct cells, return an empty list for that segment.\\
8. Do not count the same cell more than once.\\

Gold answer cells:\\
- "\{CELL\_ID\_1\}": \{HOUSE\_1\} | \{ATTRIBUTE\_1\} = \{VALUE\_1\}\\
- "\{CELL\_ID\_2\}": \{HOUSE\_2\} | \{ATTRIBUTE\_2\} = \{VALUE\_2\}\\
...\\

Reasoning segments:\\
{}[Segment 1]\\
\{SEGMENT\_1\}\\

{}[Segment 2]\\
\{SEGMENT\_2\}\\

...\\

{}[Segment 10]\\
\{SEGMENT\_10\}\\

Return JSON only with this schema:\\
\{\\
\ \ "segment\_assessments": [\\
\ \ \ \ \{\\
\ \ \ \ \ \ "segment\_idx": 1,\\
\ \ \ \ \ \ "explicitly\_correct\_cell\_ids": ["House 1.Name", "House 2.PhoneModel"],\\
\ \ \ \ \ \ "reason": "short explanation"\\
\ \ \ \ \}\\
\ \ ]\\
\}\\

The array must contain exactly 10 items, with segment\_idx = 1, 2, 3, ..., 10.
\end{orangepromptbox}

\section{Prefix Help Annotation}
\label{app:prefix_help}

To better understand what type of information enables early free-generation transfer, we conduct an LLM-assisted annotation on AIME prefixes where the receiver is incorrect without any provider prefix but first becomes correct after receiving a partial provider prefix.
For each such case, we classify the first helpful prefix according to its main role for the original question. We use a separate DeepSeek judge model (\texttt{deepseek-v4-flash})~\cite{deepseekapi}. The judge is run with temperature 0, JSON-only output, and thinking disabled.

We use three labels.
\textit{Approach-level information} indicates that the prefix mainly provides a useful setup, representation, transformation, decomposition, case distinction, equation setup, or general line of attack.
\textit{Derivation-level information} indicates that the prefix already contains concrete intermediate progress, such as a derived equation, numerical relation, constraint, intermediate result, partial computation, or substantial partial derivation.
\textit{Mixed or unclear} is used when both types are strongly present or when the prefix is insufficient to decide.

We use LLM-assisted annotation and manually inspect 50 random cases, of which 42 match our judgment.
We therefore treat the labels as coarse descriptive signals rather than definitive annotations.

\paragraph{Annotation prompt.}
The annotation model receives the original AIME question and the provider prefix snippet.
The variables \texttt{\{candidate.question\}} and \texttt{\{candidate.provider\_prefix\}} are replaced by the AIME problem and the prefix being classified, respectively.

\textbf{System prompt.}
\begin{bluepromptbox}{System prompt}
You classify short text snippets relative to a question.\\
Judge the snippet's main role for the given question.\\
Use the question and the snippet.\\
Do not restate the problem.\\
Do not solve the problem.\\
Return JSON only.
\end{bluepromptbox}

\textbf{User prompt template.}
\begin{orangepromptbox}{User prompt template}
Judge the role of the text snippet for the question below.\\
Classify the snippet using the A/B/C labels below, and give a short reason.\\

Rules:\\
1. Use the question and the [Text snippet].\\
2. Judge the snippet's role for the question, not in isolation.\\
3. Do not use any material other than the question and the text snippet.\\
4. Do not restate the problem.\\
5. Do not solve the problem.\\
6. Explain the judgment clearly.\\
7. Return JSON only.\\

{}[Question]\\
\{candidate.question\}\\

{}[Text snippet]\\
\{candidate.provider\_prefix\}\\

Return JSON with this schema:\\
\{\\
\ \ "snippet\_role\_type": "A|B|C",\\
\ \ "snippet\_role\_reason": "explanation"\\
\}\\

Label definitions:\\
A. approach-level information\\
The snippet mainly establishes how to approach the problem.\\
It may introduce a useful representation, variable setup, transformation, decomposition, case distinction, equation setup, or general line of attack.\\
The snippet does not yet contain a developed intermediate derivation or a substantial partial solution.\\
The key information is about setting up the solution path.\\

B. derivation-level information\\
The snippet has already entered the intermediate derivation stage.\\
It contains concrete progress within a solution path, such as a derived equation, constraint, numerical relation, intermediate result, partial computation, or substantial partial derivation.\\
The key information is not only how to start, but an actual derived step toward the answer.\\

C. mixed or unclear\\
Use this label if both types are strongly present and neither is clearly dominant, or if the snippet does not contain enough information to decide.
\end{orangepromptbox}

\paragraph{Qualitative examples.}
Table~\ref{tab:prefix_help_examples} shows representative examples of the annotated prefix-help types.
For readability, we show shortened problem statements and prefix snippets.

\begin{table*}[t]
\centering
\small
\resizebox{\textwidth}{!}{
\begin{tabular}{p{0.12\linewidth}p{0.16\linewidth}p{0.47\linewidth}p{0.19\linewidth}}
\toprule
Type & Case & Prefix snippet excerpt & Why this label \\
\midrule
Approach-level
& R1-Llama $\leftarrow$ GPT, Problem 3
& ``We have a game of removing 1 or 4 tokens. It's a normal play impartial combinatorial game. Determine for which \(n\) Bob ... has a forced win ... We need to compute P-positions ... For subtraction game with moves \(\{1,4\}\). Standard Nim-like ... Position \(n\) is losing if \(n-1\) and \(n-4\) are both winning ...''
& The prefix mainly sets up the solution path: it defines losing positions, gives the recurrence, and indicates that the model should compute the resulting pattern from small cases. \\

\midrule
Derivation-level
& R1-Llama $\leftarrow$ GPT, Problem 26
& ``We need count of nonnegative integer solutions to \(a+b+c=300\) ... We can factor expression: \(\sum_{\mathrm{sym}} a^2b = (a+b+c)(ab+bc+ca)-3abc\) ... Given \(a+b+c=300\). So expression \(=300(ab+bc+ca)-3abc=6{,}000{,}000\). Divide by 3: \(100(ab+bc+ca)-abc=2{,}000{,}000\).''
& The prefix has already entered a concrete derivation: it rewrites the expression using a symmetric-polynomial identity and derives an intermediate equation. \\

\midrule
Mixed or unclear
& R1-Llama $\leftarrow$ GPT, Problem 31
& ``We need to interpret numbers in base \(b\). For base \(b>9\), digits allowed 0--9, and also maybe letters for \(>9\) but digits are 1 and 7 for \(17_b\), 9 and 7 for \(97_b\).''
& The prefix is too short to classify cleanly. It suggests a possible approach, but does not contain enough substance to clearly count as either a developed setup or a derivation. \\
\bottomrule
\end{tabular}}
\caption{Representative examples for the prefix-help annotation on AIME.}
\label{tab:prefix_help_examples}
\end{table*}

\section{Complete Prefix-Accuracy Trajectories}
\label{app:all_prefix_curves}

This appendix reports the full set of prefix-accuracy trajectories for all provider--receiver pairs across datasets and transfer modes.

\begin{figure*}[t]
    \centering
    \includegraphics[width=\textwidth]{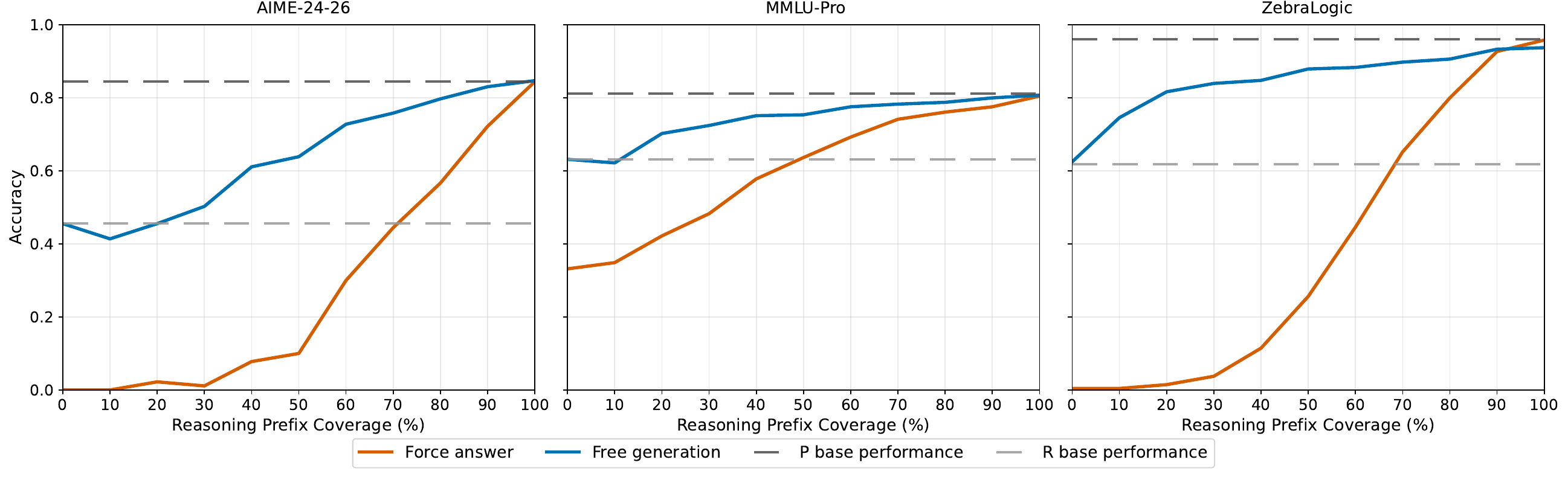}
    \caption{
Prefix-transfer trajectories for Qwen3-4B-Thinking-2507 as the provider and Qwen3-1.7B as the receiver across three datasets.
Curves show force-answer and free-generation transfer; horizontal lines mark provider and receiver base performance.
}
    \label{fig:pair_qwen4b_thk_to_qwen17b_3datasets}
\end{figure*}

\begin{figure*}[t]
    \centering
    \includegraphics[width=\textwidth]{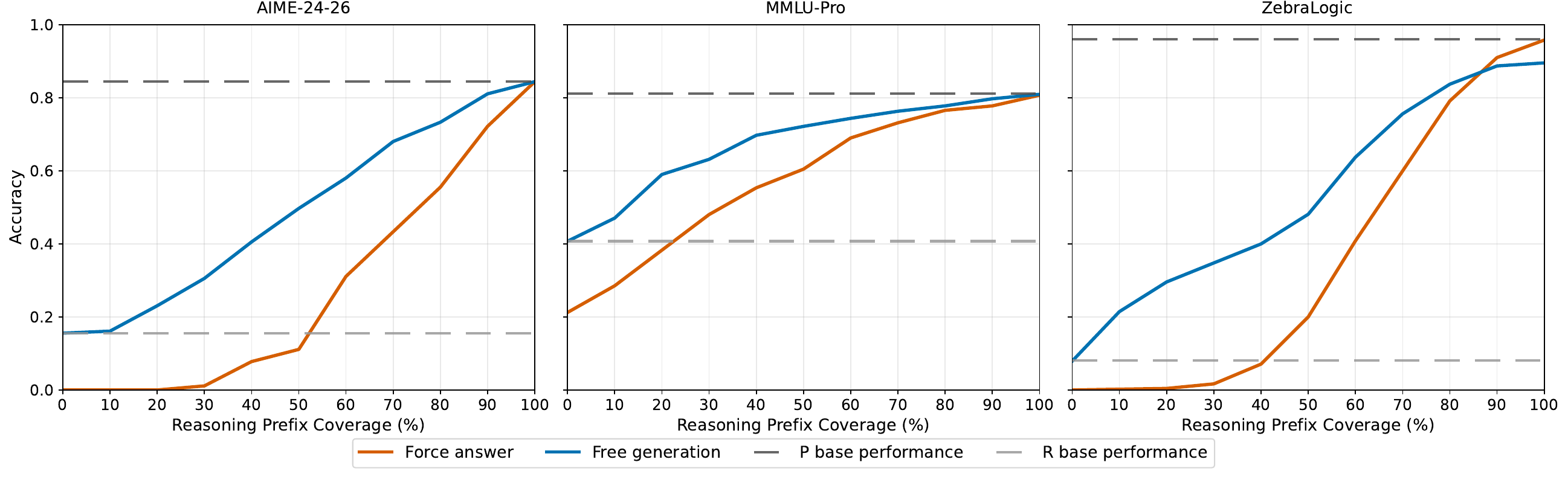}
    \caption{
Prefix-transfer trajectories for Qwen3-4B-Thinking-2507 as the provider and Qwen3-0.6B as the receiver across three datasets.
Curves show force-answer and free-generation transfer; horizontal lines mark provider and receiver base performance.
}
    \label{fig:pair_qwen4b_thk_to_qwen06b_3datasets}
\end{figure*}

\begin{figure*}[t]
    \centering
    \includegraphics[width=\textwidth]{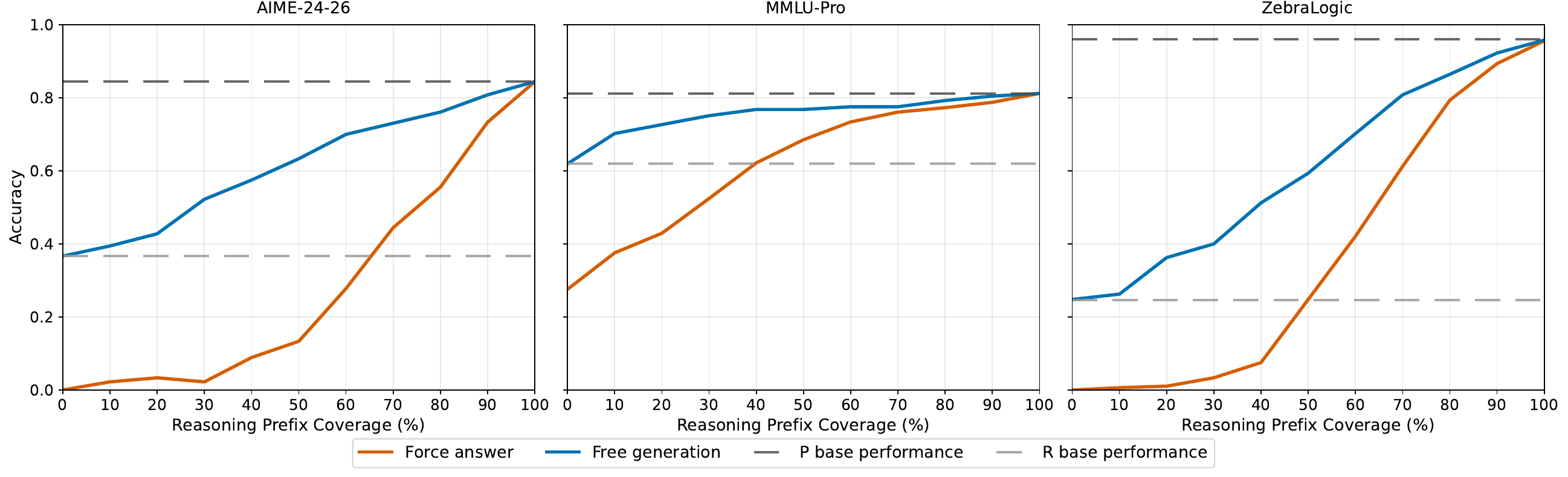}
    \caption{
Prefix-transfer trajectories for Qwen3-4B-Thinking-2507 as the provider and DeepSeek-R1-Distill-Llama-8B as the receiver across three datasets.
Curves show force-answer and free-generation transfer; horizontal lines mark provider and receiver base performance.
}
    \label{fig:pair_qwen4b_thk_to_r1llama_3datasets}
\end{figure*}

\begin{figure*}[t]
    \centering
    \includegraphics[width=\textwidth]{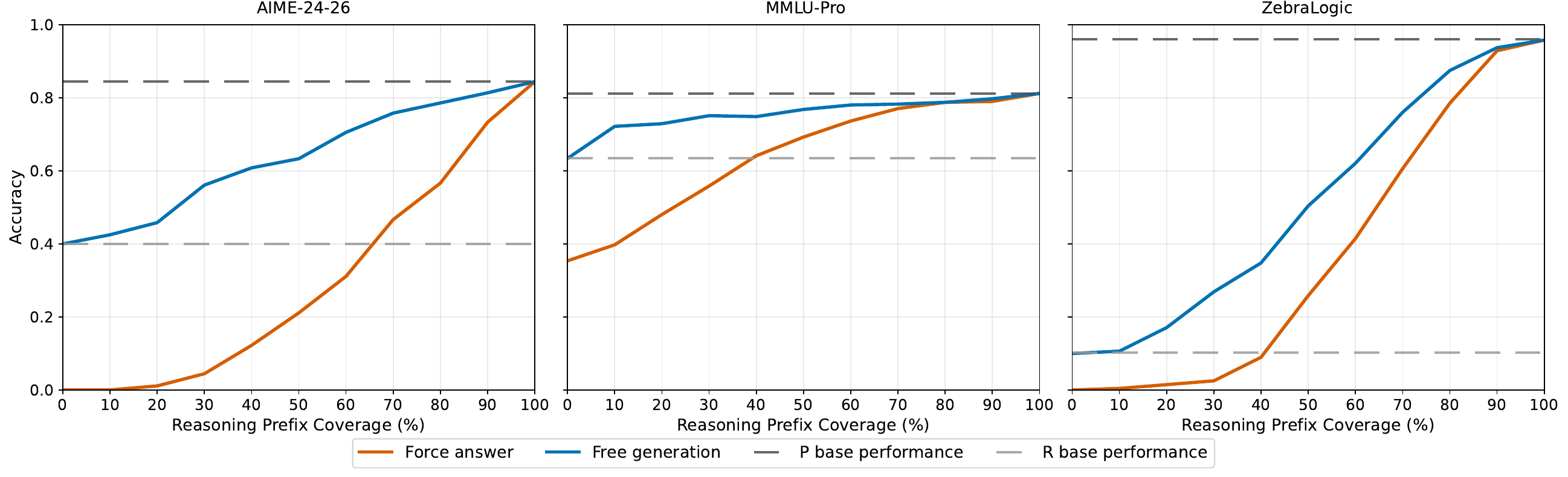}
    \caption{
Prefix-transfer trajectories for Qwen3-4B-Thinking-2507 as the provider and SmolLM3-3B as the receiver across three datasets.
Curves show force-answer and free-generation transfer; horizontal lines mark provider and receiver base performance.
}
    \label{fig:pair_qwen4b_thk_to_smollm_3datasets}
\end{figure*}

\begin{figure*}[t]
    \centering
    \includegraphics[width=\textwidth]{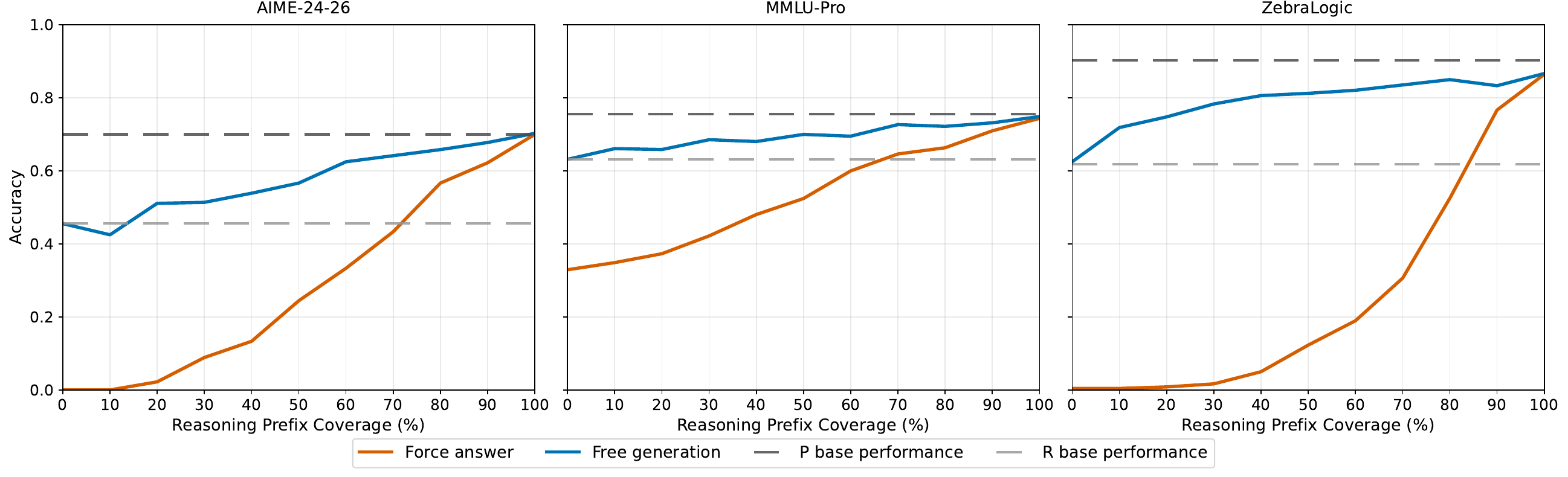}
    \caption{
Prefix-transfer trajectories for Qwen3-4B as the provider and Qwen3-1.7B as the receiver across three datasets.
Curves show force-answer and free-generation transfer; horizontal lines mark provider and receiver base performance.
}
    \label{fig:pair_qwen4b_to_qwen17b_3datasets}
\end{figure*}

\begin{figure*}[t]
    \centering
    \includegraphics[width=\textwidth]{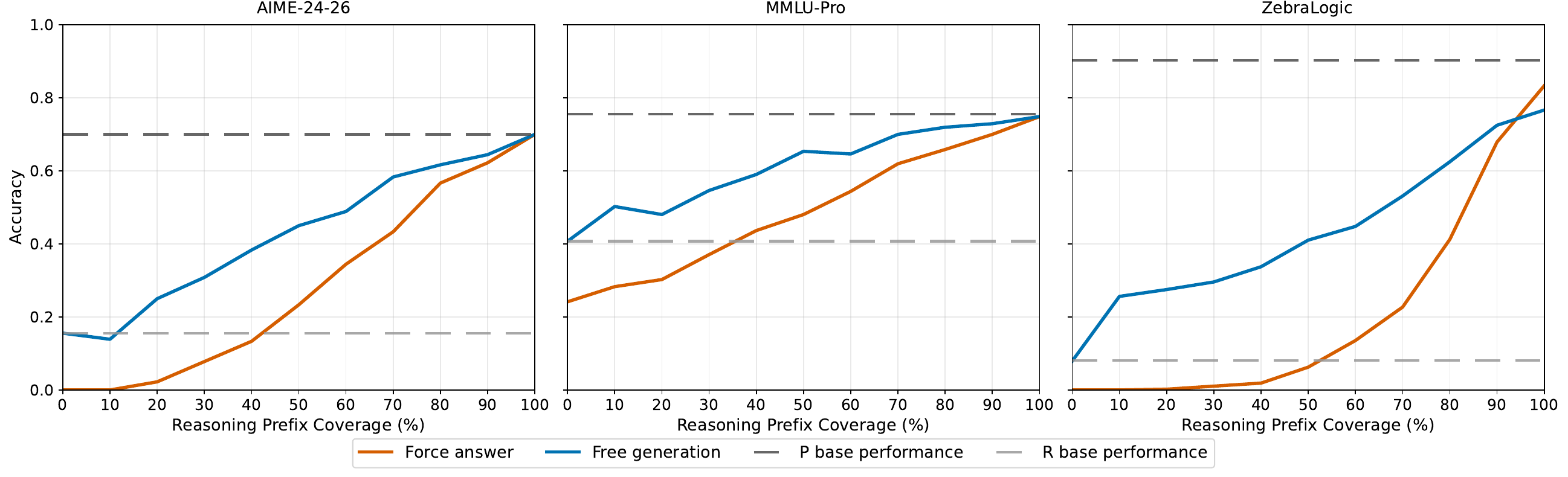}
    \caption{
Prefix-transfer trajectories for Qwen3-4B as the provider and Qwen3-0.6B as the receiver across three datasets.
Curves show force-answer and free-generation transfer; horizontal lines mark provider and receiver base performance.
}
    \label{fig:pair_qwen4b_to_qwen06b_3datasets}
\end{figure*}

\begin{figure*}[t]
    \centering
    \includegraphics[width=\textwidth]{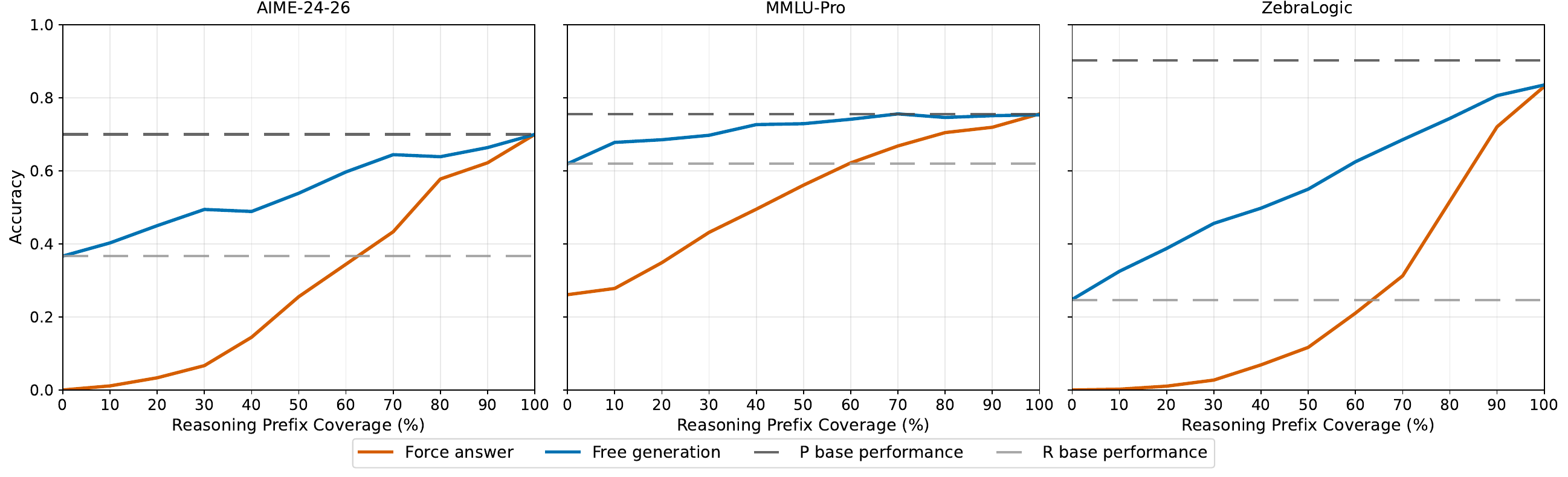}
    \caption{
Prefix-transfer trajectories for Qwen3-4B as the provider and DeepSeek-R1-Distill-Llama-8B as the receiver across three datasets.
Curves show force-answer and free-generation transfer; horizontal lines mark provider and receiver base performance.
}
    \label{fig:pair_qwen4b_to_r1llama_3datasets}
\end{figure*}

\begin{figure*}[t]
    \centering
    \includegraphics[width=\textwidth]{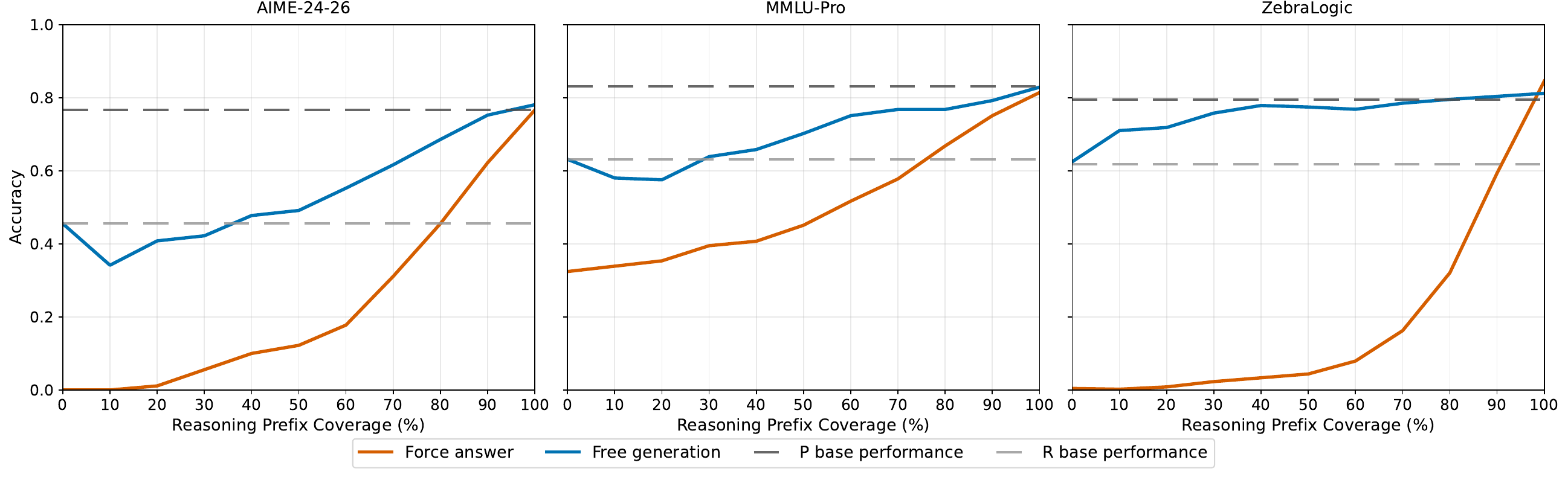}
    \caption{
Prefix-transfer trajectories for GPT-OSS-20B as the provider and Qwen3-1.7B as the receiver across three datasets.
Curves show force-answer and free-generation transfer; horizontal lines mark provider and receiver base performance.
}
    \label{fig:pair_gptoss_to_qwen17b_3datasets}
\end{figure*}

\begin{figure*}[t]
    \centering
    \includegraphics[width=\textwidth]{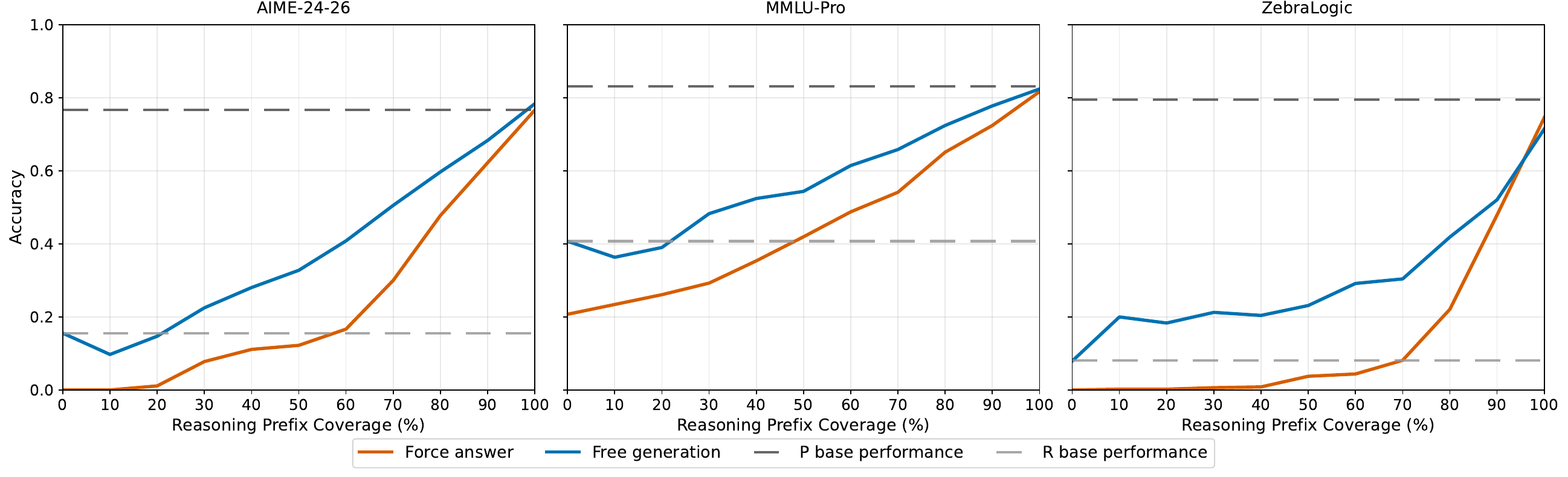}
    \caption{
Prefix-transfer trajectories for GPT-OSS-20B as the provider and Qwen3-0.6B as the receiver across three datasets.
Curves show force-answer and free-generation transfer; horizontal lines mark provider and receiver base performance.
}
    \label{fig:pair_gptoss_to_qwen06b_3datasets}
\end{figure*}

\begin{figure*}[t]
    \centering
    \includegraphics[width=\textwidth]{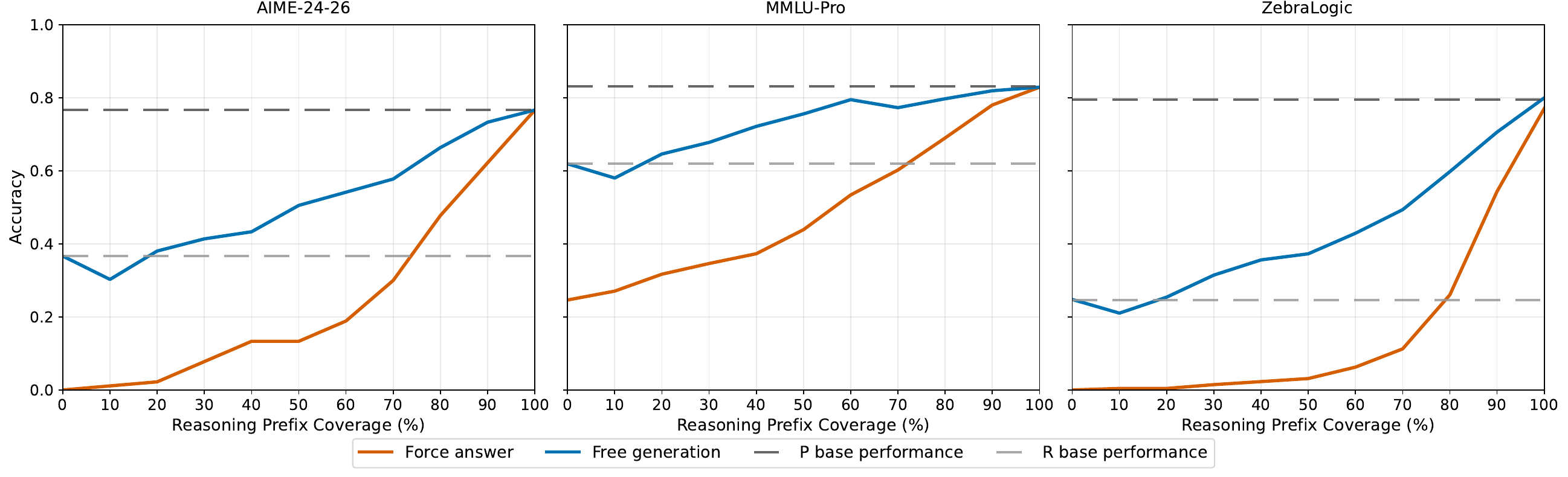}
    \caption{
Prefix-transfer trajectories for GPT-OSS-20B as the provider and DeepSeek-R1-Distill-Llama-8B as the receiver across three datasets.
Curves show force-answer and free-generation transfer; horizontal lines mark provider and receiver base performance.
}
    \label{fig:pair_gptoss_to_r1llama_3datasets}
\end{figure*}

\begin{figure*}[t]
    \centering
    \includegraphics[width=\textwidth]{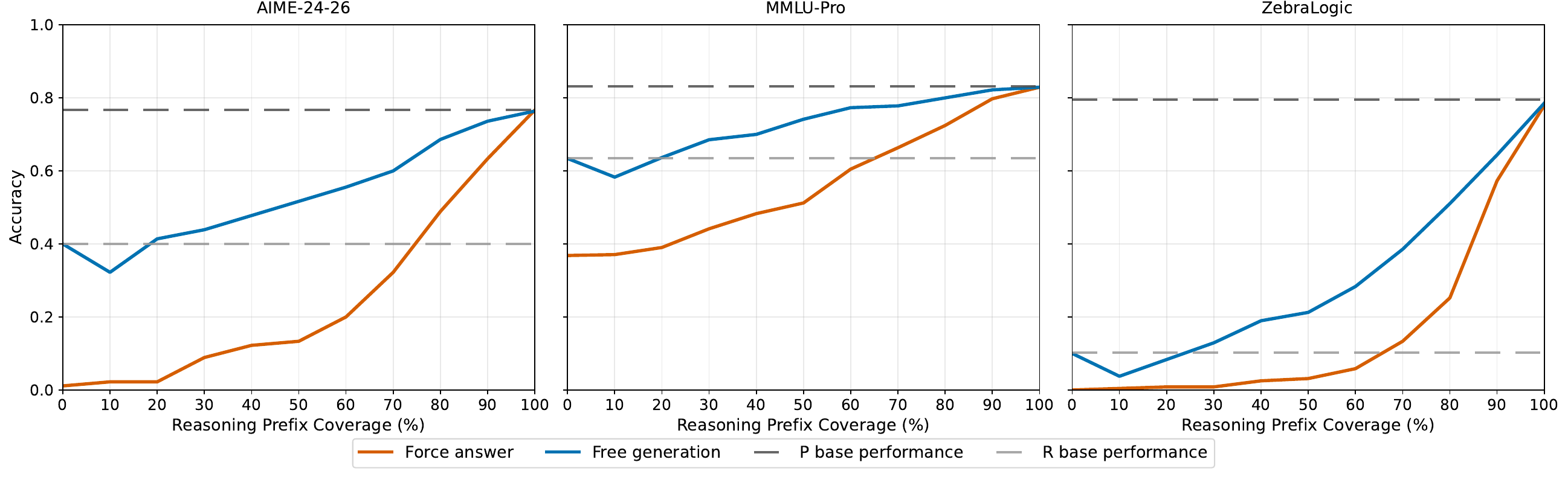}
    \caption{
Prefix-transfer trajectories for GPT-OSS-20B as the provider and SmolLM3-3B as the receiver across three datasets.
Curves show force-answer and free-generation transfer; horizontal lines mark provider and receiver base performance.
}
    \label{fig:pair_gptoss_to_smollm_3datasets}
\end{figure*}

\end{document}